\documentclass{article}
\newcommand{\tb}[1]{\textbf{#1}}
\usepackage[T1]{fontenc}     % Add this line to ensure scalable fonts
\usepackage{ragged2e}        % Provides better justification tools
\usepackage[protrusion=true,expansion=false]{microtype}  % Disable expansion
\usepackage{graphicx}
\usepackage{multirow}
\usepackage{caption}
\usepackage{float}
\usepackage{booktabs}
\usepackage{array}
\usepackage{makecell}
\usepackage{multicol}
\usepackage{tablefootnote}
\usepackage{tabularx}        % Added - for flexible table columns
\usepackage[table,xcdraw]{xcolor}
\usepackage{amsmath,amssymb,amsfonts}
\usepackage{mathtools} 
\usepackage{amsthm}
\usepackage{mathrsfs}
\usepackage{algorithm}
\usepackage{algpseudocode}  % Using algpseudocode for better formatting
\usepackage{algcompatible}% 
\usepackage{algorithmicx}

\usepackage{textcomp}
\usepackage{pifont}
\usepackage{setspace,lineno}
\usepackage{listings}
\usepackage{soul}            % For highlighting text
\sethlcolor{yellow}          % Set highlight color to yellow
\usepackage[title]{appendix}
\usepackage{manyfoot}

\usepackage{xcolor}
\definecolor{customgreen}{HTML}{00B050}
\definecolor{captionblue}{HTML}{0070C0}  

\newcommand{\cmark}{\textcolor{customgreen}{\ding{51}}} 
\newcommand{\xmark}{\textcolor{red}{\ding{55}}} 
\DeclareCaptionLabelFormat{boldblue}{\textbf{\textcolor{captionblue}{#1 #2}}}
\usepackage{titlesec}
\titleformat{\section}
  {\normalfont\Large\bfseries\color{captionblue}}{\thesection}{1em}{}
\titleformat{\subsection}
  {\normalfont\large\bfseries\color{captionblue}}{\thesubsection}{1em}{}
\titleformat{\subsubsection}
  {\normalfont\normalsize\bfseries\color{captionblue}}{\thesubsubsection}{1em}{}
\usepackage{geometry}
\usepackage{amsmath} 
\allowdisplaybreaks[4]
\usepackage[protrusion=true,expansion=false]{microtype}
\usepackage{hyperref}
\hypersetup{
    colorlinks=true,
    linkcolor=blue,
    citecolor=blue,
    urlcolor=blue
}

\title{Hallucination to Truth: A Review of Fact-Checking and Factuality Evaluation in Large Language Models}
\author{
Subhey Sadi Rahman\textsuperscript{1}, 
Md. Adnanul Islam\textsuperscript{1,†}, 
Md. Mahbub Alam\textsuperscript{1,†}, \\
Musarrat Zeba\textsuperscript{1,†},
Md. Abdur Rahman\textsuperscript{1},
Sadia Sultana Chowa\textsuperscript{2},\\
Mohaimenul Azam Khan Raiaan\textsuperscript{1,3}, 
Sami Azam\textsuperscript{3,*}\\
\small
\textsuperscript{1}Department of Computer Science and Engineering, United International University, Dhaka 1212, Bangladesh\\
\small
\textsuperscript{2} Department of Computer Science and Engineering, Daffodil International University, Dhaka-1341, Bangladesh\\
\small
\textsuperscript{3}Faculty of Science and Technology, Charles Darwin University, Casuarina, NT 0909, Australia\\
\small 
\textsuperscript{†} Equal Contributions. \\
\small 
\textsuperscript{*} Corresponding Author: sami.azam@cdu.edu.au
}
\date{} % Leave date blank
\begin{document}
\justifying
\twocolumn[
\maketitle
\begin{abstract} 
\noindent

Large Language Models (LLMs) are trained on vast and diverse internet corpora that often include inaccurate or misleading content. Consequently, LLMs can generate misinformation, making robust fact-checking essential. This review systematically analyzes how LLM-generated content is evaluated for factual accuracy by exploring key challenges such as hallucinations, dataset limitations, and the reliability of evaluation metrics. The review emphasizes the need for strong fact-checking frameworks that integrate advanced prompting strategies, domain-specific fine-tuning, and retrieval-augmented generation (RAG) methods. It proposes five research questions that guide the analysis of the recent literature from 2020 to 2025, focusing on evaluation methods and mitigation techniques. Instruction tuning, multi-agent reasoning, and RAG frameworks for external knowledge access are also reviewed. The key findings demonstrate the limitations of current metrics, the importance of validated external evidence, and the improvement of factual consistency through domain-specific customization. The review underscores the importance of building more accurate, understandable, and context-aware fact-checking. These insights contribute to the advancement of research toward more trustworthy models.

\end{abstract}

\vspace{0.5em}
\vspace{1em}
]
%\linenumbers
\section{Introduction}

The growing use of Large Language Models (LLMs) in news, healthcare, education, and law means that their accuracy directly affects real-world decisions \cite{kampelopoulos2025review, wang2025building}. These models often generate reliable information but may be false, which poses a risk of misinformation \cite{augenstein2024factuality}. As newer fact-checking techniques, datasets, and benchmarks are published rapidly, it is challenging for researchers and practitioners to keep track of effective solutions \cite{huang2025content}. This paper presents a systematic review of fact-checking related to LLMs, organizing recent advancements in this field, identifying and categorizing ongoing challenges, and highlighting potential avenues for future research. This is crucial as LLMs increasingly shape the way information is trusted and used in society.

\begin{figure*}[ht!]
    \centering
    \includegraphics[scale=0.096]{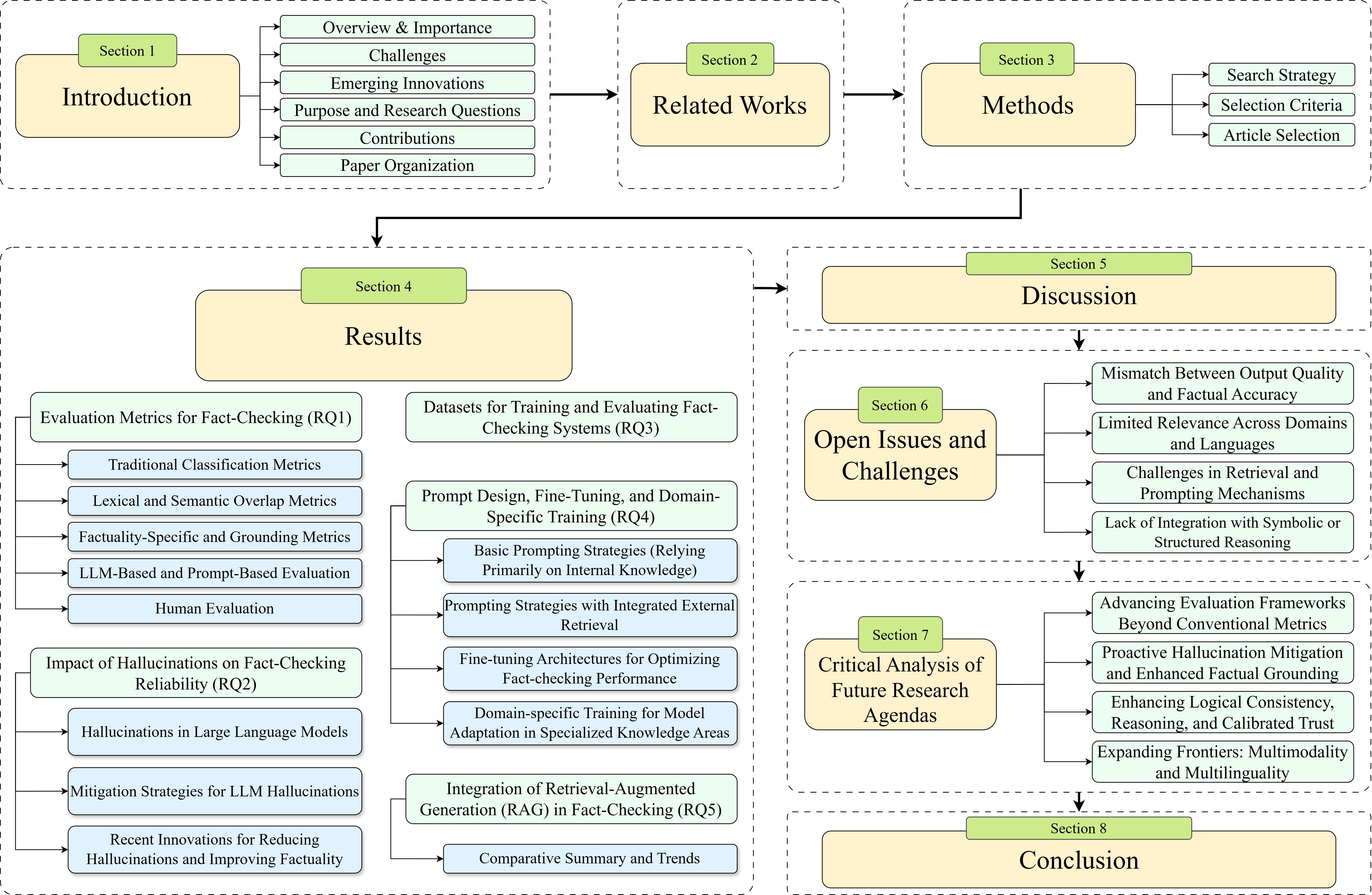}
    \caption{The fundamental content structure and categorization of this survey. The framework outlines eight sections, beginning with introduction, related works, and methods, followed by results addressing five research questions on evaluation metrics, hallucinations, datasets, prompting strategies, and retrieval-augmented generation. Subsequent sections cover discussion, open challenges, future research agendas, and conclusion, providing a coherent roadmap of the survey.}
    \label{fig:content_structure}
\end{figure*}
\subsection{Challenges}
Fact-checking the output of LLM systems faces several intimidating challenges. The lack of standardized evaluation metrics is one of the most notable challenges. Currently used ones quantify surface-level similarity rather than factual consistency \cite{lin2021truthfulqa, bender2021dangers} and are therefore less effective at detecting nuanced errors.

Another key limitation of LLMs is hallucination. They tend to produce linguistically consistent but factually inaccurate or entirely fictional text \cite{augenstein2024factuality}. This occurs due to language modeling and training on potentially stale or biased data using the probabilistic method \cite{paullada2021data, ladhak2023pre}. Dataset quality is also critical for fact-checking system performance. The majority of benchmarks either lack the realistic complexity of real-world claims and are domain-independent or are too narrow to be generalized. Furthermore, datasets with imbalanced classes can influence the model response and make the system less robust on a wide spectrum of topics \cite{aly2021feverous}.

\subsection{Emerging Innovations}
To address these issues, various innovations have been suggested, including Retrieval-Augmented Generation (RAG), instruction tuning~\cite{tang2024minicheck, setty2024surprising}, domain-specific fine-tuning~\cite{hu2024bad}, multi-agent systems~\cite{kupershtein2025ai, zhao2024pacar}, automated self-correction and feedback mechanisms~\cite{peng2023check, ma2025local}, and integration with knowledge graphs~\cite{ghosh2024logical, giarelis2024unified}. RAG stands out as a key technique that combines LLMs and external retrieval systems, aligning generated outputs with verifiable sources. RAG architectures have shown notable results in factuality and explainability~\cite{kasai2023realtime} by allowing LLMs to access and cite external knowledge in real time. These methods are often augmented with advanced prompting techniques, such as hierarchical step-by-step reasoning and multi-agent collaboration~\cite{DBLP:conf/naacl/LiPGGZ24/self-checker, zhang2023towards, khaliq2024ragar}.

\subsection{Purpose and Research Questions}
The objective of the review is to critically evaluate the current prospects of LLM-based fact-checking systems, identify key issues, and investigate the performance of existing solutions. Observing recent growing trends, this paper aims to build more accurate, transparent, and scalable fact-checking systems in LLMs. The review is inspired by five fundamental research questions (RQ).

\begin{enumerate}
    \item \textbf{RQ1:} What evaluation metrics are used to assess LLM-based fact-checking systems?\\
    \textit{Rationale:} To understand how system performance is measured and identify potential limitations or inconsistencies in current evaluation methods.
    \item \textbf{RQ2:} How do hallucinations affect the reliability of LLM fact-checking?\\
    \textit{Rationale:} Hallucinations are caused by LLM due to the vast amount of training data containing refined and unverified information. Their output often contains hallucinated answers, which directly affect the trustworthiness and accuracy of the LLM fact-checking.
    \item \textbf{RQ3:} What datasets are commonly used for training and evaluating fact-checking models?\\
    \textit{Rationale:} To assess the quality, coverage, and impact of the dataset on generalizability.
    \item \textbf{RQ4:} How do prompting strategies and fine-tuning influence fact-checking performance?\\
    \textit{Rationale:} To analyze optimization techniques for LLMs in fact-checking contexts.
    \item \textbf{RQ5:} How is RAG integrated into fact-checking?\\
    \textit{Rationale:} To evaluate the benefits and challenges of combining retrieval mechanisms with generative models.
\end{enumerate}

\subsection{Contributions}
This paper presents three contributions to research on fact-checking using LLMs: 

\vspace{0.02\linewidth}\noindent\textbf{First.} It offers a comprehensive taxonomy of evaluation metrics that categorizes widely used techniques by their methodological focus.

\vspace{0.02\linewidth}\noindent\textbf{Second.} The review combines a wide range of approaches to mitigate hallucinations in LLM output. These range from fine-tuning through domain-specific data to instruction tuning, adversarial training, and self-supervised feedback methods like Self-Checker \cite{DBLP:conf/naacl/LiPGGZ24/self-checker}. In addition, multi-agent architecture and multi-step reasoning techniques are explored to enhance factuality and explainability.

\vspace{0.02\linewidth}\noindent\textbf{Third.} The paper offers novel insight into how dataset characteristics such as domain specificity, annotation quality, and multilingual coverage affect the performance of fact-checking systems \cite{aly2021feverous}.

\subsection{Paper Organization}
This paper is organized into eight sections: Section \ref{relatedworks} reviews existing research on LLM-based fact-checking and highlights key gaps. Section \ref{methodology} explains the methodology, including how the studies were selected and analyzed. Section \ref{result_findings} presents the findings based on our five core research questions. Section \ref{discussion} discusses the implications, challenges, and limitations. Section \ref{open_issues_challenges} draws attention to open issues and challenges. Section \ref{critical_analysis_future} highlights the analysis of future research agendas, and Section \ref{conclusion} concludes the paper with key insights for building more accurate and reliable LLM-based fact-checking systems. Figure \ref{fig:content_structure} illustrates the overall structure of the paper.

\begin{table*}[ht]
\centering
\caption{Comparison of different papers concerning the key RQs. The following RQs guide our review: RQ1: Evaluation Metrics and Gaps; RQ2: Hallucination Effects and Mitigation; RQ3: Datasets and Impact; RQ4: Prompt and Fine-tuning; Domain-specific Training Effects; and RQ5: RAG and Domain-Specific Implementation Challenges.}
\label{tab:rq-coverage}
\begin{scriptsize}
\begin{tabular}{p{2.5cm}cccccp{5cm}}
    \midrule
    \textbf{Paper} 
    & \makecell{\textbf{Metrics} \\ \textbf{\& Gaps}} 
    & \makecell{\textbf{Hallucination} \\ \textbf{\& Mitigation}} 
    & \makecell{\textbf{Datasets} \\ \textbf{\& Impact}} 
    & \makecell{\textbf{Prompt, Fine-tuning} \\ \textbf{\& Domain Training}} 
    & \makecell{\textbf{RAG} \\ \textbf{\& Domain}} 
    & \textbf{Key Notes} \\
    \midrule
    Vykopal et al. \cite{vykopal2024generative} \rule{0pt}{3ex} 
    & \xmark & \cmark & \xmark & \cmark & \xmark 
    & 1. Skips hallucination effects (RQ2) \newline
    2. Only mentions domain-specific fine-tuning (RQ4) \newline
    3. Only mentions RAG's impact (RQ5) \\
    \midrule
    Dmonte et al. \cite{dmonte2025claim} \rule{0pt}{3ex} 
    & \cmark & \cmark & \cmark & \cmark & \xmark 
    & 1. Skips domain-specific datasets (RQ3) \newline
    2. Only mentions domain-specific fine-tuning (RQ4) \newline
    3. Only mentions RAG's impact (RQ5) \\
    \midrule
    Augenstein et al. \cite{augenstein2024factuality} \rule{0pt}{3ex} 
    & \cmark & \cmark & \cmark & \cmark & \xmark 
    & 1. Skips domain-specific datasets (RQ3) \newline
    2. Mentions prompt design in (RQ4) \newline
    3. Only mentions RAG's impact (RQ5) \\
    \midrule
    Wang et al. \cite{wang2024factuality} \rule{0pt}{3ex} 
    & \cmark & \cmark & \cmark & \cmark & \xmark 
    & Only mentions RAG's impact (RQ5) \\
    \midrule
    Ours \rule{0pt}{3ex} 
    & \cmark & \cmark & \cmark & \cmark & \cmark 
    & Fully addresses all RQs \\
    \midrule
\end{tabular}
\end{scriptsize}
\end{table*}

\section{Related Works}
\label{relatedworks}
LLM-generated texts are now widely used in various important sectors. Therefore, it is important to ensure their factual accuracy and reliability to maintain trust in these applications.

Several researchers have explored fact-checking methods in the context of LLMs. For example, Vykopal et al. \cite{vykopal2024generative} conduct a survey of approaches and techniques used in automated fact-checking using generative LLMs, such as claim detection, evidence retrieval, and fact verification. They introduce the concept of RAG, which can be used to mitigate challenges such as hallucinations and the use of out-of-date model knowledge, using external evidence. However, it does not address the effects of domain-specific training on LLM-based fact-checking, the challenges of RAG implementation, or how the quality of the dataset, the specificity of the domain, and the evaluation metrics influence the effectiveness of LLMs. Dmonte et al. \cite{dmonte2025claim} also explore LLM-based claim verification by analyzing full-system pipelines that include key stages such as evidence retrieval, prompt construction, and explanation generation. They review RAG techniques, including iterative retrieval and claim decomposition, which allow them to address issues such as hallucinations and the challenges of verifying complex or long claims. Evaluation metrics like FactScore and FEVER Score are highlighted to assess and improve factual accuracy. Similar to \cite{vykopal2024generative}, this paper overlooks dataset quality, domain-specific challenges, and RAG implementation issues, which are significant in evaluating the reliability of LLM-based fact-checking.

In another work, Augenstein et al. \cite{augenstein2024factuality} study the challenges of factual correctness in LLM. They focus on hallucinations, knowledge editing to reduce hallucinations, and the impact of misinformation that AI can spread, which also includes concerns about trust and misuse. They propose some mitigation strategies, such as RAG, although the discussion lacks depth on domain-specific RAG implementations and the associated challenges. Key evaluation metrics for LLM-based fact-checking systems are also discussed to assess factuality, consistency, and text quality, including TruthfulQA, FactScore, GPTScore, G-Eval, SelfCheckGPT, BERTScore, and MoverScore. However, the paper lacks a discussion of several critical technical aspects and challenges, including model interpretability, explainability, and practical implementation and integration of the proposed mitigation techniques in real-world settings. Wang et al. \cite{wang2024factuality} offer a detailed survey of the factuality of LLM, providing a taxonomy of hallucination types and errors in both unimodal and multimodal tasks. A key contribution is mapping factuality challenges to algorithmic solutions and proposing improvements to factuality-aware model calibration. However, the paper does not discuss domain-specific challenges of RAG, prompt design, fact-checking component integration, or system-level architectures and deployment strategies for LLM verification environments.

Although most previous work has briefly discussed evaluation metrics, hallucination effects, and issues related to prompt- or fine-tuning methods, it has not examined the impact of datasets, particularly domain-specific ones. In addition, there is a wide gap between practical challenges and considerations of RAG and domain-specific implementation. Most of the papers give RAG only a passing reference without discussing larger domain-specific problems. However, our work thoroughly addresses the primary research questions, especially the under-discussed areas, and thus provides a more detailed overview of the field. Table \ref{tab:rq-coverage} presents a summary of the existing survey papers and shows how they relate to the key research questions of this study.

\section{Methods}
\label{methodology}
To explore how LLMs can be applied to fact-checking, we adopted a structured and practical approach inspired by well-established research methods \cite{DBLP:journals/infsof/KitchenhamBBTBL09}. The review process consisted of three key phases: (i) planning, (ii) data collection and analysis, and (iii) synthesis and reporting. \tb{Firstly, we defined the overall scope of the review and ensured alignment with our RQs (i.e., RQ1-RQ5). This included developing a detailed review protocol that specified the objectives, inclusion and exclusion criteria, and the databases to be searched. We also identified the dimensions most relevant to LLM fact-checking, including evaluation metrics, hallucinations, datasets, prompting and fine-tuning methods, and RAG, and set these as the guiding categories for further analysis.} 

We then performed a broad search across leading academic databases, using a combination of manual screening and automated tools to identify the most relevant and up-to-date studies. \tb{Following this, we have analyzed the studies according to the methodological approach employed (i.e., benchmarking, prompting strategy, dataset evaluation), which RQ it addressed, and the contribution type (e.g., framework, metric, dataset, or application).} At the end of our review, we brought together \tb{these} insights from the selected studies to highlight what is already known, where the gaps are, and what future research should aim to address. By following this clear and step-by-step methodology, we have ensured that our findings are informative and reliable to advance the use of LLM in fact-checking tasks \cite{conway1968committees}. \tb{The following sections detail the process in each aspect.}

{ %
\subsection{Search Strategy}
To ensure a comprehensive review of the relevant literature, we developed a focused search strategy employing well-defined keywords and Boolean operators. Our objective was to capture publications related to the application of LLMs in fact-checking and associated tasks. The search was conducted across several major academic databases, including Web of Science, arXiv, Scopus, and OpenReview, and covered a publication window from January 1, 2021, to September 15, 2025.
For the Scopus and Web of Science databases, we utilized the following comprehensive search query:

\begin{quote}
\small
("large language models" AND "fact-checking") OR ("LLM" AND "misinformation detection") OR ("automated fact verification" AND "LLMs") OR ("factuality evaluation" AND "natural language processing") OR ("hallucination" AND "LLMs") OR ("LLM hallucination") OR ("hallucination mitigation" AND "large language models") OR ("hallucination detection" AND "large language models") OR ("fact-checking datasets" OR "benchmark datasets for fact verification") OR ("retrieval-augmented generation" AND "fact-checking") OR ("RAG" AND "LLM") OR ("RAG" AND "fact-checking") OR ("fine-tuning" AND "fact verification models") OR ("prompt engineering" AND ("truthful generation" OR "fact-checking")) OR ("LLM-based fact verification" AND "NLP") OR (("misinformation detection" OR "fake news") AND "hallucination")
\end{quote}

The specific search keys used for OpenReview and arXiv are detailed in our publicly available GitHub repository. Our selection process was designed to target significant research areas, including fact-checking methodologies, model assessment techniques, dataset analysis, and optimization procedures.
The initial search produced the results summarized in Table \ref{tab:search_results}.

\begin{table}[h!]
\centering
\caption{Number of Publications Retrieved from Each Database}
\label{tab:search_results}
\begin{scriptsize}
\begin{tabular}{lr}
\toprule
\textbf{Database} & \textbf{Total Papers Retrieved} \\
\midrule
Web of Science    & 1,235                           \\
arXiv             & 9,180                           \\
Scopus            & 3,270                           \\
OpenReview        & \textasciitilde1,000            \\
\midrule
\textbf{Total}    & \textbf{14,685}                 \\
\bottomrule
\end{tabular}
\end{scriptsize}
\end{table}

The high number of results from arXiv is primarily a consequence of its search limitations. The platform does not handle complex, multi-term queries well, which requires us to search for each keyword separately. This method inevitably led to the same articles being counted multiple times, explaining the inflated total.
}

\subsection{Selection Criteria}
\begin{figure*}[ht!]
    \centering
    \includegraphics[scale=0.37]{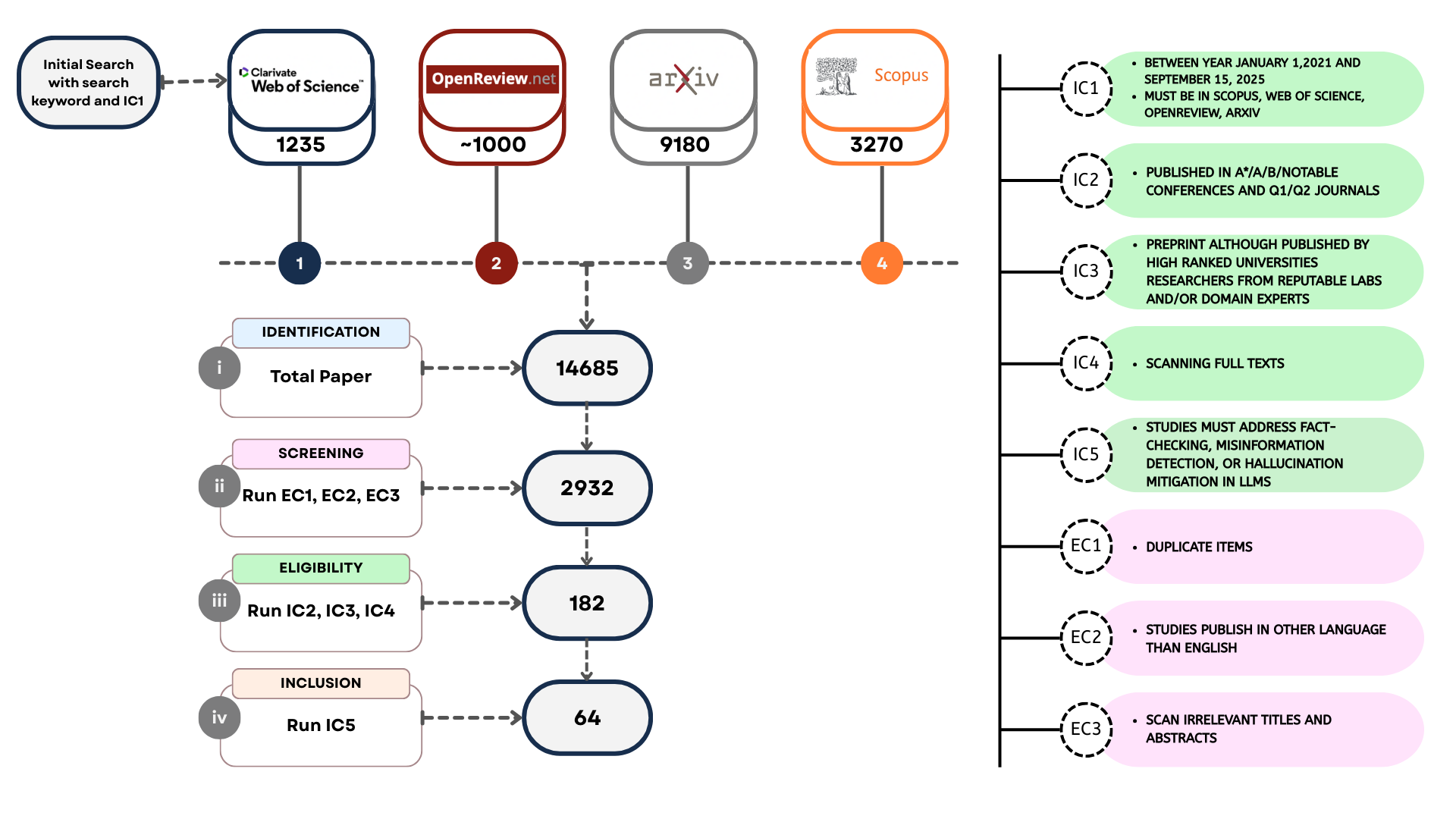}
    \caption{The article selection and screening process is on the left, and inclusion and exclusion criteria for article selection are on the right.}
    \label{fig:ArticleSelection_and_InclusionExclusion}
\end{figure*}

Each article is carefully assessed using our evaluation criteria to determine whether it meets the inclusion or exclusion requirements. The key inclusion criteria (IC) and the exclusion criteria (EC) are shown in Figure \ref{fig:ArticleSelection_and_InclusionExclusion}.

\subsection{Article Selection} 
We conducted a systematic review of articles from leading conferences and journals at the intersection of fact-checking, Natural Language Processing (NLP), and LLMs. The review focused on studies published between January 1, 2021, and September 15, 2025, that employed LLMs to verify external claims or factual content, excluding works that solely analyzed hallucinations or internal factual consistency.

From an initial pool of 14,685 records retrieved from various academic databases, we applied a multi-stage screening process, comprising duplicate removal, title and abstract screening, and full-text evaluation. The article selection workflow is illustrated in Figure \ref{fig:ArticleSelection_and_InclusionExclusion}, which outlines the stages of identification, screening, eligibility, and final inclusion. Finally, we selected 64 articles that meet our inclusion criteria and align with the objectives of this review, with publication dates ranging from September 2022 to August 2025.

\begin{figure}[ht!]
\centering
\includegraphics[scale=0.11]{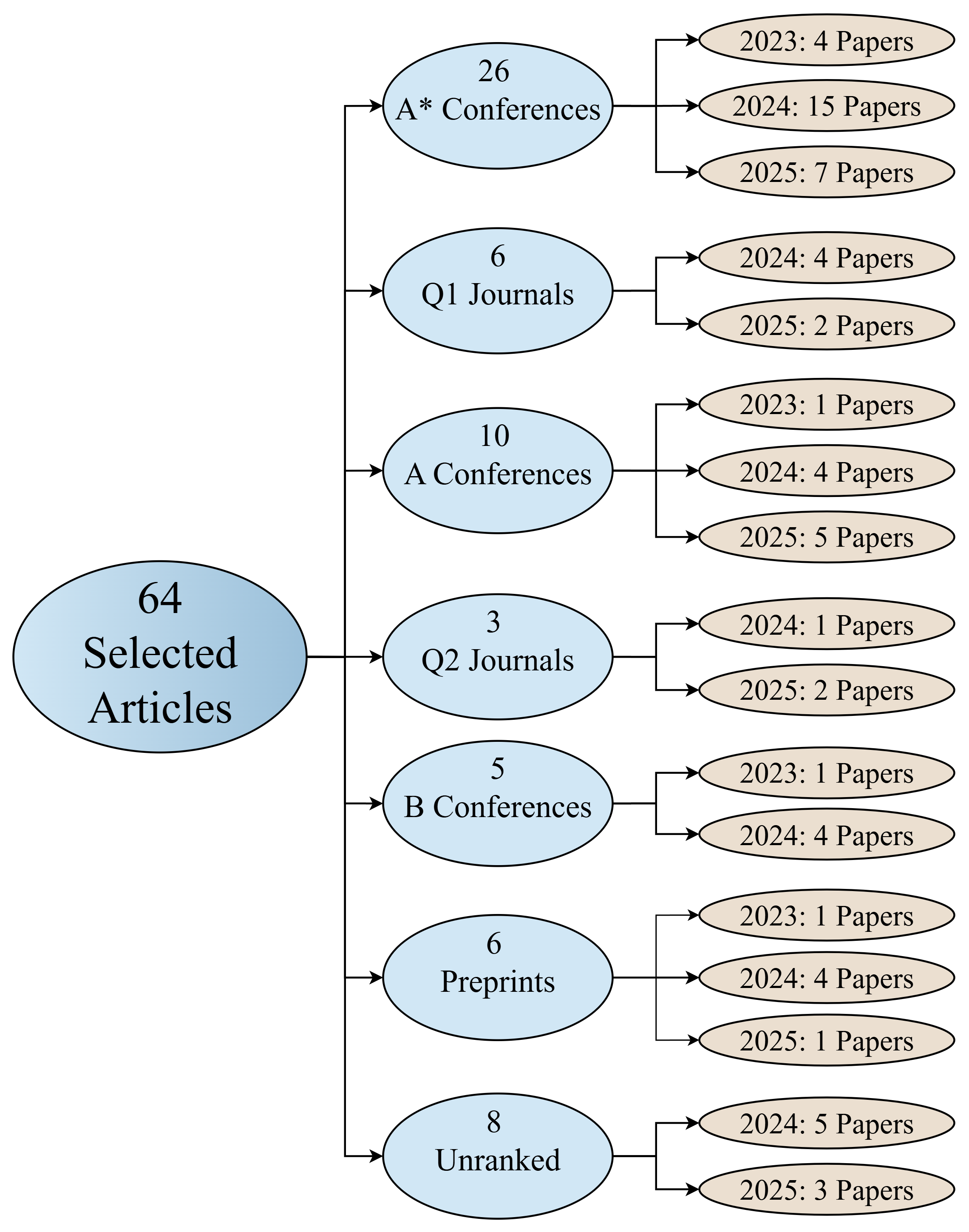}
\caption{A visual summary of the articles selected from journals, conference proceedings, and preprints.}
\label{fig:Selected_Article_Summary}
\end{figure}

\begin{figure}[ht!]
\centering
\includegraphics[scale=0.2]{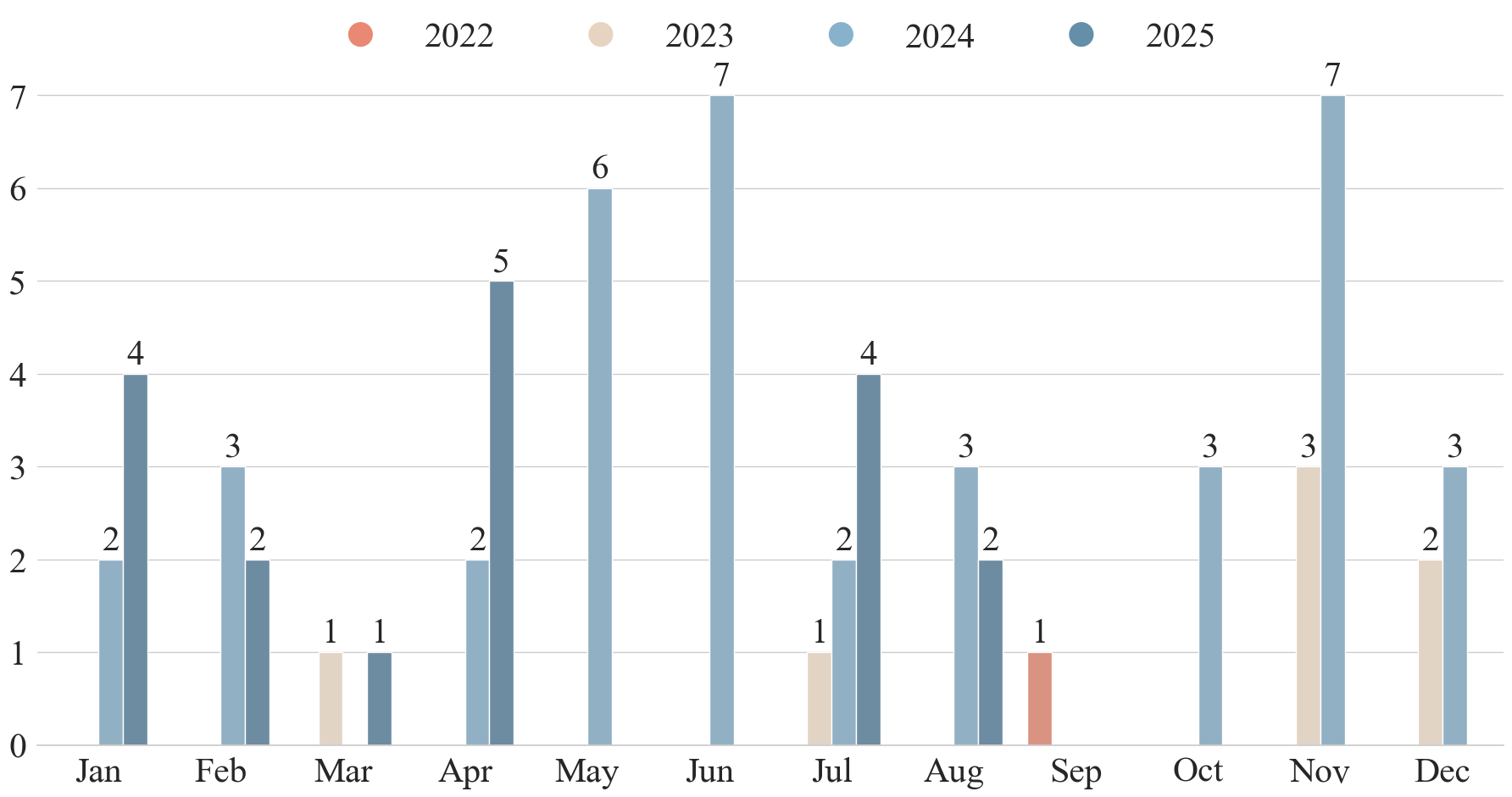}
\caption{Grouped bar chart showing the monthly breakdown of publications from 2021 to 2025.}
\label{fig:Monthly_Distribution_ArticlesJournals}
\end{figure}

\begin{figure}[ht!]
\centering
\includegraphics[scale=0.2]{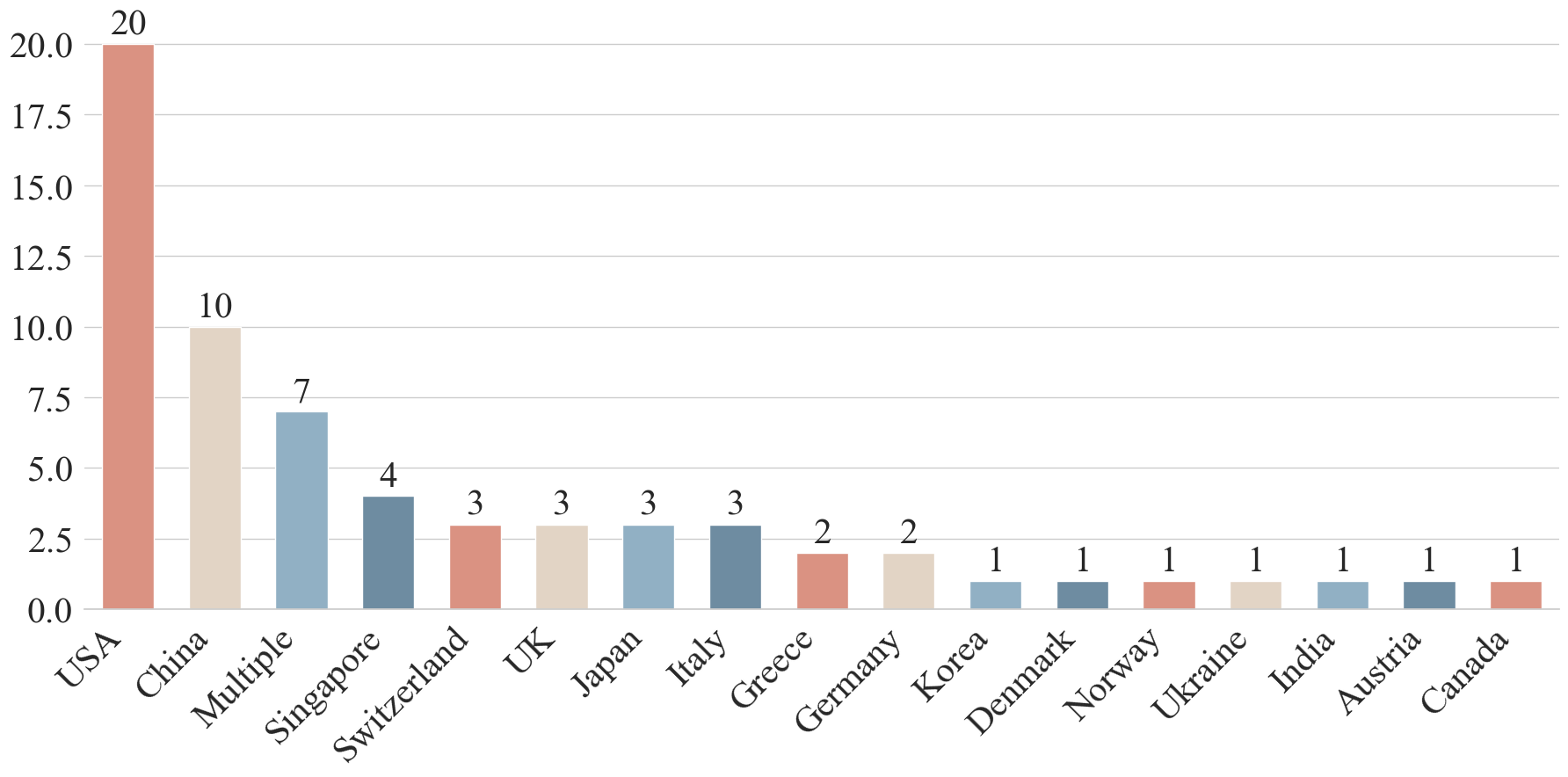}
\caption{Bar chart illustrating the geographical breakdown of publications.}
\label{fig:Geo_Distribution_ArticlesJournals}
\end{figure}

An overview of the number of selected journals, conference proceedings, and preprints is shown in Figure \ref{fig:Selected_Article_Summary}. Figure \ref{fig:Monthly_Distribution_ArticlesJournals} demonstrates the monthly publication frequency of the selected articles, with the majority published in 2024 and 2025. Furthermore, Figure \ref{fig:Geo_Distribution_ArticlesJournals} summarizes the geographical distribution of the publications, indicating that regions with well-established research ecosystems and institutional support contributed the largest share of works, which often correlates with higher representation in prestigious venues and greater citation visibility.

To ensure a comprehensive scope, we examined the distribution of publication venues. Due to the fact that various channels influence methodological priorities, venue analysis is crucial. RAG, prompting techniques, and assessment measures are commonly highlighted at prestigious NLP conferences, including ACL, EMNLP, and NeurIPS. Understanding publication frequency in conjunction with venue dispersion sheds light on the field's structural and temporal dynamics. Because it has a direct impact on methodological variety and research impact, venue concentration is important. Peer review procedures at high-impact conferences and publications are usually more stringent, guaranteeing methodological originality, more robust empirical validation, and wider recognition. Because of this, work that is published in these types of forums typically receives more attention and citations, which supports the prevailing paradigms in the area. This pattern also reflects how research environments with greater institutional support and global visibility tend to secure stronger representation in prestigious venues, which in turn shapes methodological diversity and amplifies citation impact. Domain-specific journals, on the other hand, provide important diversity by publishing contextually grounded research that may not meet the innovation-focused requirements of prestigious venues, despite the fact that they are frequently less referenced.

% \begin{figure*}[ht!]
%     \centering
%     \includegraphics[scale=0.12]{Figures/Paper_selection.drawio.png}
%     \caption{The article selection and screening process of this survey.}
%     \label{fig:paper_selection}
% \end{figure*}

\section{Findings from the Research Questions}
\label{result_findings}
In this section, we present a comprehensive key result finding focusing on evaluation metrics, the impact of hallucinations on LLMs, datasets, prompt design and fine-tuning, and integration of RAG.

\subsection{Evaluation Metrics for Fact-Checking Systems (RQ1)}

\begin{figure*}[ht!]
    \centering
    \includegraphics[scale=0.13]{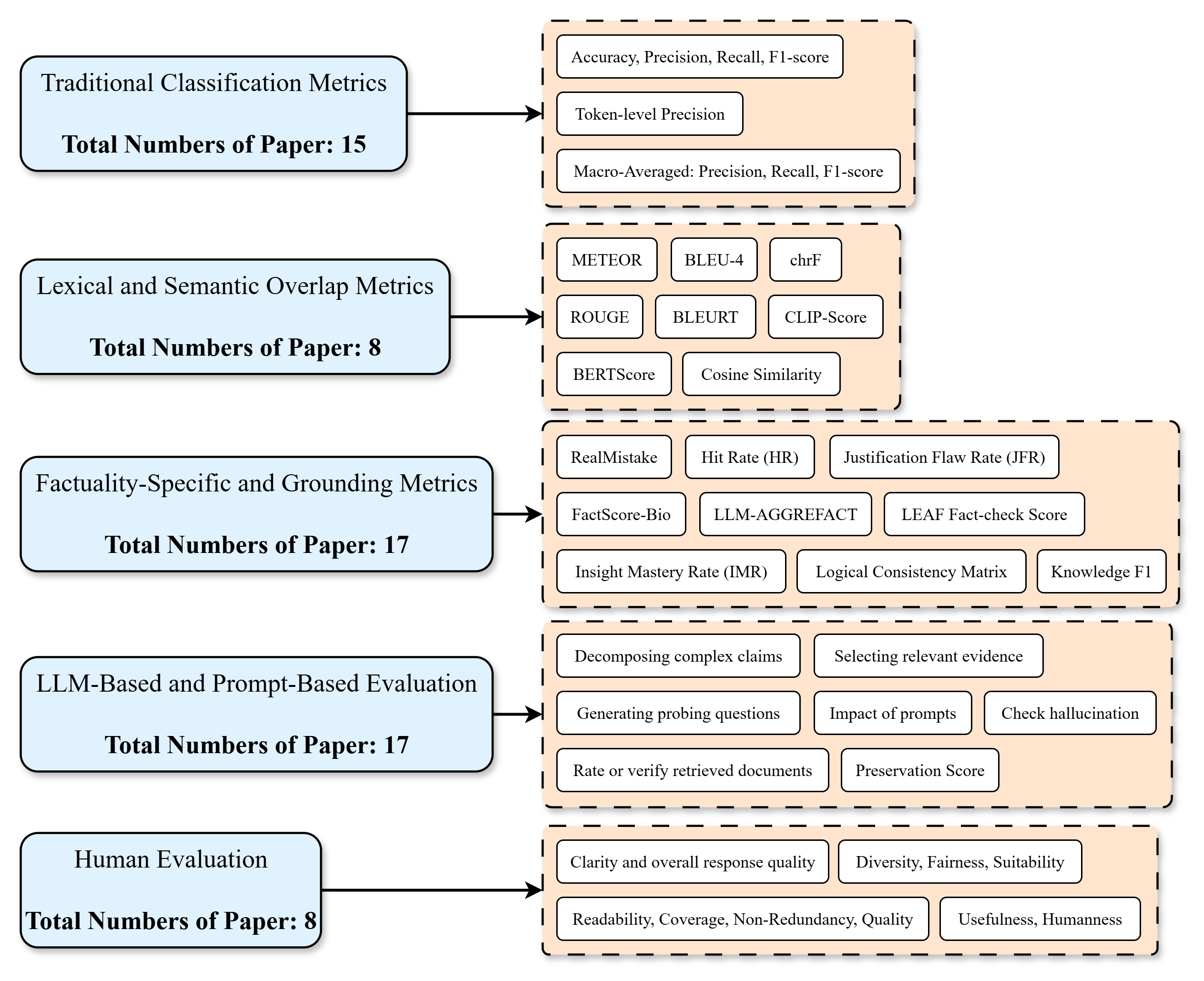}
    \caption{Taxonomy of evaluation metrics for fact-checking systems. The framework categorizes metrics into five groups. Each category is illustrated with representative measures and the number of papers adopting them, highlighting the breadth of methodologies employed in fact-checking research.}
    \label{fig:section - 4.1}
\end{figure*}

Evaluating LLMs for fact-checking and related areas such as grounded generation, summarization, and error detection is a crucial and evolving field. It addresses one of the most significant challenges in LLM deployment: their tendency to “hallucinate,” or generate text that sounds plausible but is factually incorrect \cite{tang2024minicheck, giarelis2024unified}. Evaluation in this context typically involves checking the model’s output against provided evidence or reliable external sources \cite{tang2024minicheck}. Previously, a wide range of methods have been developed, including using LLMs themselves as evaluators and building benchmarks that unify datasets and tasks \cite{tang2024minicheck}. Figure \ref{fig:section - 4.1} illustrates a comprehensive summary of the complete set of metrics.

\subsubsection{Traditional Classification Metrics}

Most often, evaluation tasks are approached as classification problems, which determine whether a claim is true or identify errors in responses \cite{zhang2025dataset, kamoi2024evaluating}. Metrics like accuracy \cite{zhang2025dataset,tran2024leaf, qi2024sniffer, singhal2024multilingual, si-etal-2024-large, xie2024fire}, precision, recall, and F1-score \cite{ peng2023check, zhang2023towards,lee2020language, cao2024multi,  liu2025bidev, vladika2025step} are widely used for these tasks. In multiclass scenarios, such as classifying statements as supported, refuted, or inconclusive, macro-averaged versions of these metrics are employed \cite{ zhang2023towards, zhang2025dataset,lee2020language}. These metrics also serve as standard measures in detection tasks \cite{sankararaman2024provenance}. For short-form responses, token-level precision with annotated answers is typical \cite{peng2023check}. Token-level responses allow for partial evaluation, where an answer can be mostly correct, but as it is highly dependent on the specific tokenizer utilized, direct comparisons between different models are difficult.

These metrics offer quantitative performance indicators, making it easier to compare models or methods directly \cite{zhang2023towards,  zhang2025dataset, tran2024leaf, singhal2024multilingual}. They often reduce complex output to a binary (i.e., correct or incorrect) judgment, overlooking reasoning quality or nuanced inaccuracies \cite{giarelis2024unified}. They may also be misleading in datasets with imbalanced labels \cite{sankararaman2024provenance}.

\subsubsection{Lexical and Semantic Overlap Metrics}

When evaluating text generation tasks, such as summarization or dialogue, overlap-based metrics are commonly used. Lexical overlap metrics such as BLEU-4, METEOR, and chrF assess surface-level similarity \cite{peng2023check}. ROUGE evaluates the extent to which summaries or explanations capture the core content \cite{qi2024sniffer, magomere2025claims}. Semantic similarity metrics like BERTScore \cite{augenstein2024factuality, peng2023check,  xie2024fire, pisarevskaya2025zero}, BLEURT \cite{peng2023check}, and cosine similarity measures \cite{krishnamurthy2024yours} assess deeper semantics. For multimodal outputs, CLIP-Score compares image and caption embeddings to evaluate alignment \cite{ge2024visual}.

These metrics are standard for assessing fluency and content similarity, and can capture meaning beyond exact word matches \cite{ peng2023check, krishnamurthy2024yours, ge2024visual}. However, they do not measure the factual correctness. High overlap or semantic scores may still correspond to factually incorrect content. Older semantic metrics may not align well with the reasoning capabilities of modern LLMs \cite{augenstein2024factuality, magomere2025claims, pisarevskaya2025zero}.

\subsubsection{Factuality-Specific and Grounding Metrics}

Specialized metrics have been developed to directly evaluate the factual consistency. These go beyond surface similarity and focus on whether the model’s claims align with the evidence. For example, benchmarks like LLM-AGGREFACT use detailed human annotations to assess support levels for claims \cite{tang2024minicheck}. It is a meta-benchmark for factuality that aggregates 11 different publicly available fact-checking and hallucination datasets. ReaLMistake focuses on binary error detection, especially in reasoning and context alignment \cite{kamoi2024evaluating}. Other tools like the LEAF fact-check Score compute the ratio of factually supported sentences to the total response \cite{tran2024leaf} by decomposing each response into individual sentences, then each sentence is independently verified against retrieved external knowledge sources. Knowledge F1 (KF1), utilized by Peng et al. \cite{peng2023check}, measures the overlap between human-used and model-used knowledge. FactScore-Bio classifies responses based on retrieved evidence \cite{wang2025openfactcheck}, and some methods aggregate multiple signals into a final factuality probability score \cite{sankararaman2024provenance}. Metrics such as Insight Mastery Rate (IMR) and Justification Flaw Rate (JFR) assess explanatory quality \cite{lin2025fact} where IMR represents the proportion of low-scoring fact-checking responses relative to the total number of questions, where a Grade of three or below (on a ten-point scale) indicates errors in the target LLM’s response. On the other hand, JFR denotes the percentage of cases where the target LLM conducted correct verdict prediction yet had poor justification, based on the conditions set by IMR. Natural Language Inference (NLI) techniques and textual entailment tasks also serve to classify claims as supported, refuted, or unverifiable \cite{setty2024surprising, si-etal-2024-large, lee2020language, cheung2023factllama, choi2024automated}.

Challenges include the complexity of strict entailment in language \cite{quelle2024perils}, potential metric bias \cite{augenstein2024factuality}, and reliance on high-quality annotated evidence. The Logical Consistency Matrix measures coherence under logical manipulations like negation or conjunction \cite{ghosh2024logical, xie2024fire}. Hit Rate (HR), used in evidence retrieval, tracks how often relevant documents are among the top results \cite{zhao2024medico}. These methods are tailored for evaluating truthfulness and provide nuanced insights into factual grounding \cite{tang2024minicheck, kamoi2024evaluating, tran2024leaf, wang2025openfactcheck}.

\subsubsection{LLM-Based and Prompt-Based Evaluation}

A growing trend involves using LLMs themselves as evaluators. This includes having LLMs classify responses as correct or flawed, and rate the factual accuracy of claims when prompted \cite{tang2024minicheck, kamoi2024evaluating}. The LLM-as-a-judge paradigm treats powerful language models as referees that compare and score the outputs of other models, which often produces results that closely align with human judgments \cite{lin2025fact, jing2024scale}. LLMs are also widely used for tasks such as decomposing complex claims \cite{tang2024minicheck, krishnamurthy2024yours, zhang2024reinforcement}, generating probing questions \cite{ tran2024leaf,zhang2024reinforcement, schlichtkrull2023averitec, singhal2024evidence}, and selecting relevant evidence \cite{DBLP:conf/naacl/LiPGGZ24/self-checker, tran2024leaf, krishnamurthy2024yours, schlichtkrull2023averitec}. While techniques like zero-shot, few-shot, Chain-of-Thought, ReAct, and HiSS are not metrics themselves, their impact is assessed using factuality metrics \cite{hu2024bad, zhang2023towards, singhal2024multilingual, chatrath2024fact}. Some systems even use LLMs to rate and verify retrieved documents \cite{zhang2024reinforcement}, or use them to check for hallucinations \cite{peng2023check}. The Preservation Score evaluates how much original content remains intact after hallucination correction \cite{zhao2024medico}. LLMs enable more nuanced and context-sensitive evaluations than traditional metrics \cite{leite2023detecting}. They can reduce human effort in evaluation tasks as well \cite{lin2025fact}. Their performance can be inconsistent due to sensitivity to prompt phrasing, and they may introduce bias or misjudgments \cite{kamoi2024evaluating, leite2023detecting}.

\subsubsection{Human Evaluation}

Despite automation advances, human evaluation remains essential, especially for complex and subjective aspects like explanation clarity and overall response quality \cite{ding2025citations}. Evaluators often use Likert scales to rate Readability, Coverage, Non-Redundancy, and Quality \cite{giarelis2024unified, zhang2023towards, lin2025fact}. In dialogue tasks, several studies also assess aspects like Usefulness and Humanness \cite{peng2023check}. Human-annotated data often forms the ground truth for many benchmarks \cite{tang2024minicheck, kamoi2024evaluating, wang2025openfactcheck}, and evaluation criteria may include Redundancy, Diversity, Fairness, and Suitability \cite{lin2025fact}. Human judgments remain the gold standard, particularly for evaluating factual correctness and nuanced generation quality \cite{ giarelis2024unified, kamoi2024evaluating}, even though it is time-consuming, costly, and can introduce subjective variance depending on the evaluators and the criteria used \cite{kamoi2024evaluating, lin2025fact}.

\subsubsection{Comparative Summary and Trends}

The landscape of LLM evaluation is becoming increasingly sophisticated. Traditional metrics like Accuracy and F1-score still serve as foundational tools for classification tasks \cite{lee2020language, zhang2023towards, zhang2025dataset}. However, more advanced evaluations focused on factuality and grounding are gaining prominence, especially in response to challenges such as hallucinations \cite{tang2024minicheck,  kamoi2024evaluating, tran2024leaf, wang2025openfactcheck}.

Human-annotated benchmarks and specialized metrics help ensure robustness, while LLMs are now frequently integrated into the evaluation loop, whether to score, verify, or generate intermediate outputs \cite{tang2024minicheck, kamoi2024evaluating, singhal2024multilingual}. Although promising, these LLM-based evaluations require careful validation against human judgments due to reliability concerns \cite{kamoi2024evaluating, leite2023detecting}. Human evaluation continues to play a vital role, particularly in high-stakes and qualitative tasks. The trend is toward hybrid frameworks that combine multiple evaluation strategies (e.g., automated metrics, LLM reasoning, and human oversight) to assess LLMs more holistically \cite{peng2023check,DBLP:conf/naacl/LiPGGZ24/self-checker,  zhang2024reinforcement}. Thus, evaluating LLM fact-checking across diverse languages and modalities, including multimodal or cross-lingual fact-checking, is an emerging frontier and demands the adaptation or creation of new evaluation techniques\cite{khaliq2024ragar, singhal2024multilingual, quelle2024perils, geng2024multimodal}.

\begin{figure}[!ht]
\centering
\includegraphics[scale=0.15]{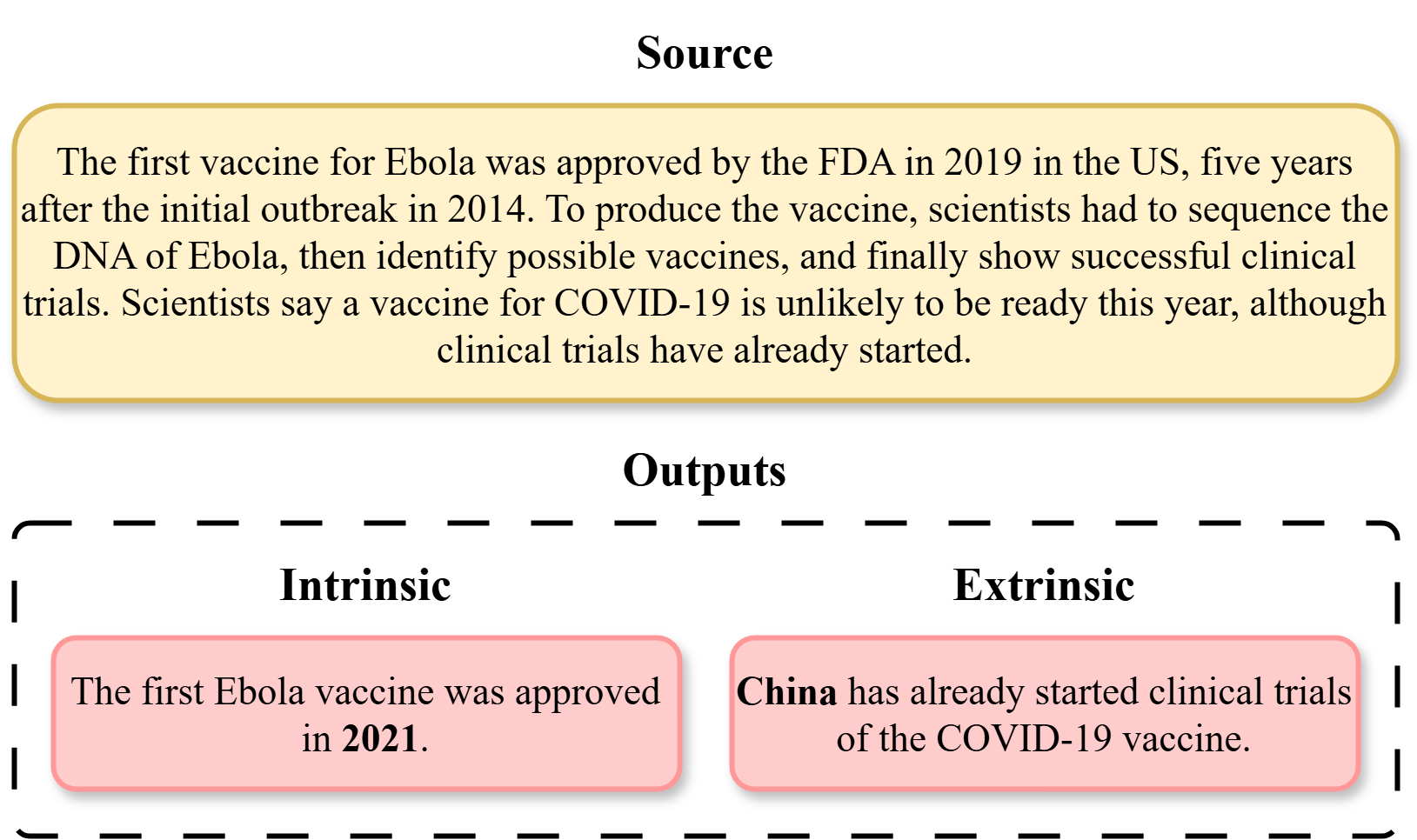}
\caption{Intrinsic vs. extrinsic hallucinations in LLM outputs: The source text provides verifiable ground truth about Ebola and COVID-19 vaccines. The intrinsic hallucination example contradicts the fact explicitly stated in the source, whereas extrinsic hallucination introduces new information that is not supported by the source.}
\label{fig:section - 4.2_in_ex}
\end{figure}

\subsection{Impact of Hallucinations on Fact-Checking Reliability (RQ2)}

Hallucinations in LLM refer to results that seem fluent, coherent, and linguistically correct but are factually inaccurate, nonsensical, unsupported, or completely fabricated \cite{bang2023multitask, guerreiro2023hallucinations}. As these outputs are presented with the same level of confidence and linguistic fluency as factually accurate statements (see Figure \ref{fig:section - 4.2_in_ex}), it is often difficult for users to detect them without external verification \cite{bang2023multitask, huang2025survey}. These hallucinations can arise from contradictory, outdated, or misleading information in training corpora, as well as biases, lack of grounding, and even from prompts \cite{augenstein2024factuality, zhang2023towards,  sakurai2024llm}. 

An overview of all the papers referenced in this section is presented in Table \ref{tab:hallucination_summary_table}.

\begin{table*}[ht!]
\centering
\caption{An overview of studies on hallucinations covered in RQ2. The table organizes prior research by types and causes of hallucinations, their implications for reliability in fact-checking, and mitigation strategies. For each category, the publication years, representative authors, and study counts are reported.}
\begin{scriptsize}
\begin{tabular}{p{5cm}p{1.2cm}p{10cm}c}
\midrule
\textbf{Topics Covered} & \textbf{Years} & \textbf{Authors} & \textbf{Count} \\
\midrule
Nature and Types of Hallucinations & 2021, 2022, 2023, 2025 & Huang et al. \cite{huang2021factual}, Ji et al. \cite{ji2023survey}, Li et al. \cite{li2022faithfulness}, Jing et al. \cite{jing2024scale}, Huang et al. \cite{huang2025survey}, Augenstein \cite{augenstein2024factuality}, Zhao et al. \cite{zhao2024medico} & 7 \\
\midrule
Causes of Hallucination & 2021, 2022, 2023, 2024, 2025 & Xie et al. \cite{xie2024fire}, Zhou et al. \cite{zhou2023lima}, Wang et al. \cite{wang2023aligning}, Peng et al. \cite{peng2023check}, Ghosh et al. \cite{ghosh2024logical}, Augenstein et al. \cite{augenstein2024factuality}, Lin et al. \cite{lin2021truthfulqa}, Bender et al. \cite{bender2021dangers}, Paullada et al. \cite{paullada2021data}, Ladhak et al. \cite{ladhak2023pre}, Weidinger et al. \cite{weidinger2021ethical}, Wang et al. \cite{wang2025openfactcheck}, Kasai et al. \cite{kasai2023realtime}, Tran et al. \cite{tran2024leaf}, Cheung et al. \cite{cheung2023factllama}, Li et al. \cite{li2022large}, Onoe et al. \cite{onoe2022entity}, Tang et al. \cite{tang2024minicheck}, Yao et al. \cite{yao2023llm} & 19 \\
\midrule
Implications for Reliability in Fact-Checking & 2023, 2024, 2025 & Peng et al. \cite{peng2023check}, 
Si et al. \cite{si-etal-2024-large}, 
Li et al. \cite{li2024large}, 
Zhao et al. \cite{zhao2024medico}, 
Xie et al. \cite{xie2024fire}, 
Quelle et al. \cite{quelle2024perils}, 
Hu et al. \cite{hu2024bad}, 
DeVerna et al. \cite{deverna2024fact}, 
Singhal et al. \cite{singhal2024evidence}, 
Zhao et al. \cite{zhao2024pacar}, 
Augenstein et al. \cite{augenstein2024factuality}, 
Wang et al. \cite{wang2025openfactcheck}, 
Jing et al. \cite{jing2024scale} & 13 \\
\midrule
Fine-tuning and Instruction Tuning & 2021, 2023, 2024, 2025 & Tang et al. \cite{tang2024minicheck}, 
Setty et al. \cite{setty2024surprising}, 
Hu et al. \cite{hu2024bad}, 
Zhang et al. \cite{zhang2025dataset}, 
Zhao et al. \cite{zhao2024pacar}, 
Cheung et al. \cite{cheung2023factllama}, 
Tran et al. \cite{tran2024leaf}, 
Qi et al. \cite{qi2024sniffer}, 
Luo et al. \cite{luo2021newsclippings}, 
Leite et al. \cite{leite2023detecting}, 
Jing et al. \cite{jing2024scale} & 11 \\
\midrule
RAG & 2023, 2024, 2025 & Singhal et al. \cite{singhal2024evidence}, 
Quelle et al. \cite{quelle2024perils}, 
Augenstein et al. \cite{augenstein2024factuality}, 
Si et al. \cite{si-etal-2024-large}, 
Peng et al. \cite{peng2023check}, 
Sankararaman et al. \cite{sankararaman2024provenance}, 
Khaliq et al. \cite{khaliq2024ragar}, 
Zhang et al. \cite{zhang2024reinforcement}, 
Xie et al. \cite{xie2024fire}, 
Tran et al. \cite{tran2024leaf}, 
Wei et al. \cite{Wei2024longform}, 
Zhao et al. \cite{zhao2024pacar}, 
Qi et al. \cite{qi2024sniffer}, 
Li et al. \cite{DBLP:conf/naacl/LiPGGZ24/self-checker}, 
Giarelis et al. \cite{giarelis2024unified}, 
Ma et al. \cite{ma2025local}, 
Zhao et al. \cite{zhao2024medico}, 
Li et al. \cite{li2024large}, 
Ghosh et al. \cite{ghosh2024logical}, 
Tang et al. \cite{tang2024minicheck} & 20 \\
\midrule
Adversarial Tuning & 2025 & Leippold et al. \cite{leippold2025automated} & 1 \\
\midrule
Automated Feedback Mechanisms and Self-Correction & 2023, 2024, 2025 & Peng et al. \cite{peng2023check}, 
Ma et al. \cite{ma2025local}, 
Xie et al. \cite{xie2024fire}, 
Tran et al. \cite{tran2024leaf}, 
Fadeeva et al. \cite{fadeeva2024fact}, 
Ghosh et al. \cite{ghosh2024logical}, 
Ge et al. \cite{ge2024visual}, 
Zhao et al. \cite{zhao2024medico} & 8 \\
\midrule
Hybrid Approaches and Multi-Agent Systems & 2023, 2024, 2025 & Zhang et al. \cite{zhang2023towards}, 
Zhao et al. \cite{zhao2024pacar}, 
Ma et al. \cite{ma2025local}, 
Kupershtein et al. \cite{kupershtein2025ai}, 
Li et al. \cite{li2024large}, 
Giarelis et al. \cite{giarelis2024unified}, 
Hu et al. \cite{hu2024bad}, 
Ghosh et al. \cite{ghosh2024logical}, 
Jing et al. \cite{jing2024scale} & 9 \\
\midrule
Multimodal Fact-Checking & 2024 & Cao et al. \cite{cao2024multi}, 
Ge et al. \cite{ge2024visual}, 
Yao et al. \cite{yao2023end},
Papadopoulos et al. \cite{papadopoulos2025red},
Sharma et al. \cite{sharma2024vision},
Qi et al. \cite{qi2024sniffer},
Wang et al. \cite{wang2024llm},
Kakizaki et al. \cite{Kakizaki25maft},
Geng et al. \cite{geng2024multimodal}, 
Khaliq et al. \cite{khaliq2024ragar} & 10 \\
\midrule
Multilingual Fact-Checking & 2024 & Quelle et al. \cite{quelle2024perils},
Zhang et al. \cite{zhang2023dont},
Shafayat et al. \cite{shafayat2024multifact},
Siino et al. \cite{siino2022fake}, 
Shcharbakova et al. \cite{shcharbakova2025scale},
Jannah et al. \cite{jannah2025multilingual} & 6 \\
\midrule
Domain-Specific Fact-Checking & 2023, 2024, 2025 & Zhang et al. \cite{zhang2025dataset}, 
Tran et al. \cite{tran2024leaf}, 
Vladika et al. \cite{vladika2025step}, 
Zhao et al. \cite{zhao2024medico}, 
Xiong et al. \cite{Xiong2024Aug}, 
Jing et al. \cite{jing2024scale}, 
Chatrath et al. \cite{chatrath2024fact}, 
Khaliq et al. \cite{khaliq2024ragar}, 
Choi et al. \cite{choi2024automated}, 
Zhang et al. \cite{zhang2023towards}, 
Hu et al. \cite{hu2024bad}, 
Leite et al. \cite{leite2023detecting}, 
Wang et al. \cite{wang2025openfactcheck}, 
Qi et al. \cite{qi2024sniffer}, 
Choi et al. \cite{choi2024fact}, 
Pisarevskaya et al. \cite{pisarevskaya2025zero}, 
Liu et al. \cite{liu2025bidev} & 17 \\
\midrule
Enhancing Explainability and Trust & 2023, 2024, 2025 & Ding et al. \cite{ding2025citations}, 
Sankararaman et al. \cite{sankararaman2024provenance}, 
Quelle et al. \cite{quelle2024perils}, 
Zhao et al. \cite{zhao2024pacar}, 
Qi et al. \cite{qi2024sniffer}, 
Vladika et al. \cite{vladika2025step}, 
Krishnamurthy et al. \cite{krishnamurthy2024yours}, 
Ghosh et al. \cite{ghosh2024logical}, 
Leite et al. \cite{leite2023detecting}, 
Giarelis et al. \cite{giarelis2024unified} & 10 \\
\midrule
Hierarchical Prompting and Multi-Step Reasoning & 2023, 2024 & Zhang et al. \cite{zhang2023towards}, 
Khaliq et al. \cite{khaliq2024ragar}, 
Zhao et al. \cite{zhao2024pacar} & 3 \\
\midrule
\end{tabular}
\end{scriptsize}
\label{tab:hallucination_summary_table}
\end{table*}

\subsubsection{Hallucinations in LLMs}
\vspace{0.02\linewidth}\noindent\textbf{\textit{Nature and Types of Hallucinations.}} In the context of fact-checking, hallucinations manifest as intrinsic or extrinsic errors \cite{huang2021factual, ji2023survey, li2022faithfulness}. Intrinsic hallucinations occur when the model’s generated output contradicts the source content. In contrast, extrinsic hallucinations introduce information that cannot be verified by any provided evidence, resulting in output unsupported by the source and often containing fabricated details \cite{jing2024scale, huang2025survey, ji2023survey}. 

\begin{figure}[!ht]
    \centering
    \includegraphics[scale=0.15]{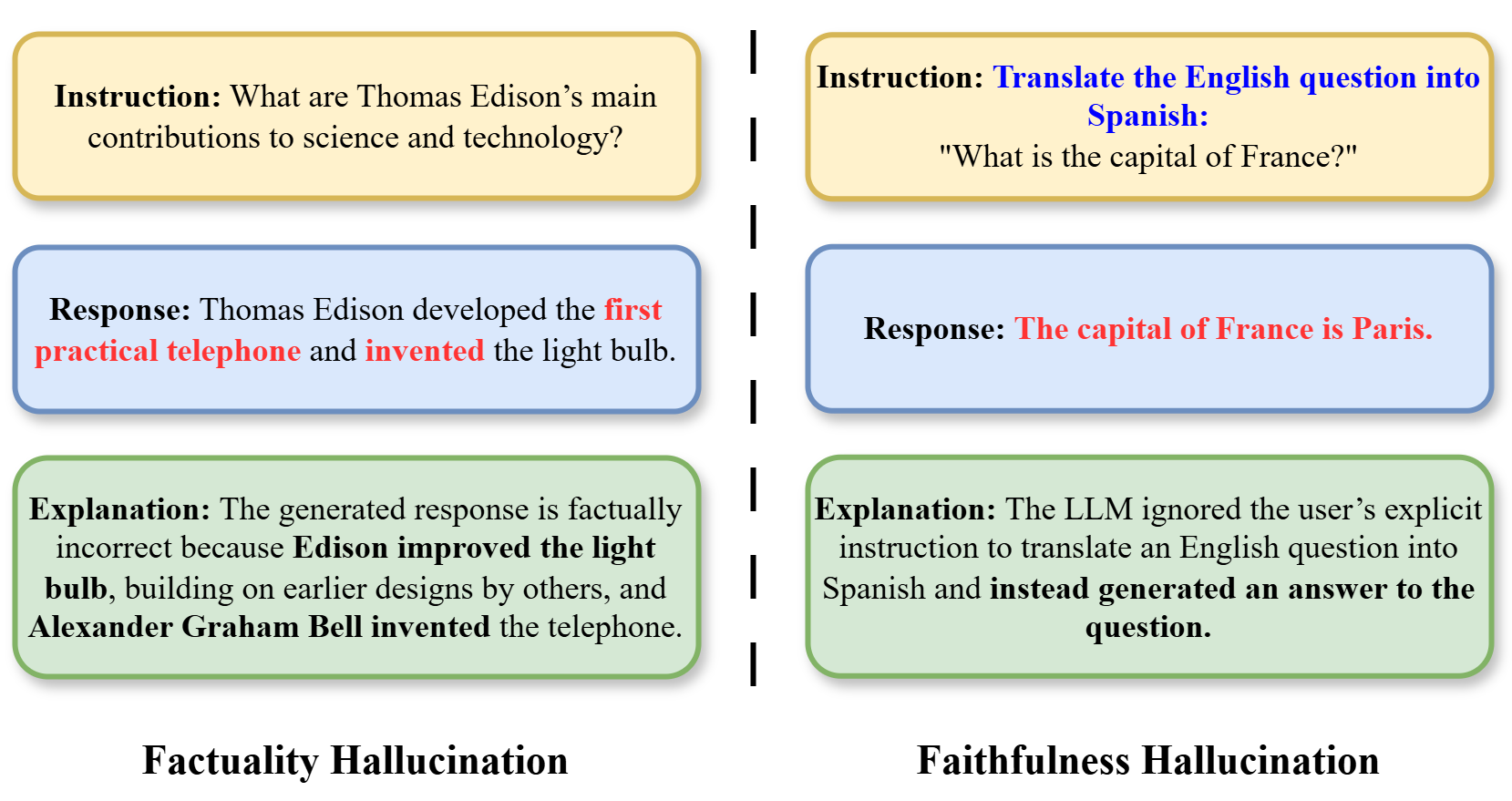}
    \caption{Two types of hallucination: \textcolor{red}{Red}-highlighted text shows hallucinated content, while \textcolor{blue}{blue}-highlighted text reflects user instructions or context that conflict with the hallucination.}
    \label{fig:section - 4.2_fact_faith}
\end{figure}

Summarization models may intrinsically hallucinate by stating a fact that contradicts the source article, or an open-ended question and answer (Q\&A) model may extrinsically hallucinate entirely new (false) information that it presents as factual. The authors of \cite{augenstein2024factuality, zhao2024medico,  huang2025survey} demonstrated that LLM hallucinations can generally be classified into two types (see example in Figure \ref{fig:section - 4.2_fact_faith}): (i) faithfulness, where the generated text does not accurately reflect the input context; and (ii) factuality, where the generated text is factually incorrect according to real-world knowledge.

\vspace{0.02\linewidth}\noindent\textbf{\textit{Causes of Hallucination.}} Hallucinations arise from fundamental misalignment in the way LLMs are trained and used. The core training objective of most LLMs is to predict the next word in a sentence based on patterns learned from massive text data, not to guarantee truthfulness \cite{xie2024fire, zhou2023lima, wang2023aligning}. This means that models are optimized to produce text that is coherent and contextually appropriate rather than factually accurate \cite{peng2023check, ghosh2024logical}. Thus, LLMs' tendency to hallucinate can be traced to their optimization focus on linguistic fluency and coherence, rather than factual precision, especially when faced with queries outside of their training distribution or when internal knowledge conflicts arise \cite{augenstein2024factuality}. During training, they absorb a large number of statements, including inaccuracies and fictional content, without an explicit mechanism to distinguish truth from falsehood \cite{lin2021truthfulqa, bender2021dangers, paullada2021data, ladhak2023pre, weidinger2021ethical}. As a result, an LLM may generate claims that sound confident and align with linguistic patterns in its memory, but are not grounded in facts. Wang et al. \cite{wang2025openfactcheck} demonstrate that a "knowledge error" can occur when the model produces hallucinated or inaccurate information due to a lack of relevant knowledge or internalizing false knowledge in the pretraining stage or the problematic alignment process.

Furthermore, since the LLM’s parametric knowledge base is fixed after training, it can become outdated or insufficient, leading to guesses on topics that it does not know \cite{kasai2023realtime, tran2024leaf, cheung2023factllama, li2022large, onoe2022entity}. When prompted “closed-book” (without access to external data), a model faced with an unknown fact will often fill the gap by generating a plausible answer, which is a major source of hallucination \cite{tang2024minicheck}. In other cases, even if grounding documents are provided, the model might improperly blend or misattribute information from these sources, causing intrinsic hallucinations \cite{tang2024minicheck}. Hallucinations may also be an inherent adversarial vulnerability of LLMs. Even nonsense or out-of-distribution prompts can trigger the model to produce a false but fluent response \cite{yao2023llm}. This suggests that, beyond knowledge gaps, the model’s sensitivity to input perturbations and over-reliance on spurious correlations in training data can induce hallucinations.

\vspace{0.02\linewidth}\noindent\textbf{\textit{Implications for Reliability in Fact-Checking.}} Hallucinations directly undermine the reliability of LLMs as fact-checkers across domains, and in some cases, they can be fatal when deployed for mission-critical tasks \cite{peng2023check, si-etal-2024-large,zhao2024medico, li2024large}. Studies estimate that even advanced models like GPT-4 and LLaMA-2 produce inaccurate factual statements in approximately 5–10\% of their responses to general knowledge queries \cite{xie2024fire}. When LLMs are used in fact-checking systems, hallucinations can lead to misclassification of claims, which can lead to an underestimation of the accuracy and trustworthiness of fact-checking procedures \cite{quelle2024perils}. Moreover, hallucinations risk amplifying misinformation when LLM-generated content aligns with existing false narratives, unknowingly facilitating their widespread acceptance in the eyes of the public \cite{hu2024bad, singhal2024evidence,deverna2024fact}. 

The complexity of detecting hallucinated content, particularly when intertwined with accurate information, complicates the verification process and increases the mental workload for human fact checkers, potentially affecting decision-making, as evidenced by studies that indicate reduced discernment after LLM-assisted fact-checking \cite{si-etal-2024-large, deverna2024fact}. Additionally, automated fact-checking systems comprising LLMs for evidence retrieval, claim interpretation, or verdict generation are highly vulnerable, as hallucinations at any stage can contaminate the entire fact-checking pipeline, ultimately resulting in unreliable outcomes \cite{augenstein2024factuality, zhao2024pacar,  wang2025openfactcheck}. Recent work emphasizes the need to quantify these phenomena systematically to enhance model reliability, proposing methodologies specifically aimed at measuring and addressing hallucination severity within faithfulness evaluations \cite{jing2024scale}. Singhal et al. \cite{singhal2024evidence} state that, in terms of fact-checking, simply assigning a veracity label is inadequate. The prediction must be supported by evidence to ensure the system’s transparency and to bolster public trust.

\subsubsection{Mitigation Strategies for LLM Hallucinations}

\vspace{0.02\linewidth}\noindent\textbf{\textit{Fine-tuning and Instruction Tuning.}} Fine-tuning pre-trained LLMs on datasets tailored to a specific domain or task, emphasizing factuality, substantially enhances their reliability \cite{tang2024minicheck}. Quantitative comparisons supporting this observation are summarized in Table \ref{tab:quantitative_summary}. Table \ref{tab:frameworksforhallucinationmitigation} illustrates different fact-checking frameworks and their demonstrated approach to mitigate hallucination. Instruction tuning, a specialized variant of fine-tuning, trains models to better adhere to explicit instructions aimed at factual and verifiable responses. Specifically, domain-specific adaptation involves fine-tuning on datasets rich in factual claims and evidence from specialized areas, such as news, medical, political, religious, or legal domains, thus helping models learn the intricacies of factual language and reasoning pertinent to these fields \cite{setty2024surprising}. Tang et al. \cite{tang2024minicheck} showed that fine-tuning transformer-based models, RoBERTa, DeBERTa, and T5 models on synthetic and ANLI data boosts robustness and improves performance. Hu et al. \cite{hu2024bad} proposed that utilizing LLMs' ability to provide rationales could enhance fine-tuned small language models (SLMs), thereby improving fake news detection performance. Zhang et al. \cite{zhang2025dataset} demonstrated that in clinical claim evaluation, domain-tuned discriminative models such as BioBERT with 80.2\% accuracy outperformed both zero-shot and fine-tuned generative LLMs like Llama3-70B, even after tuning. Zhao et al. \cite{zhao2024pacar} fine-tuned Pretrained LMs utilizing various strategies such as BERT-FC with dual loss, LIST5 with list-wise reasoning, and T5 language model, RoBERTa using NLI, and MULTIVERS with multitask learning.

\begin{table*}[ht!]
\centering
\fontsize{8}{9.6}\selectfont
\caption{An overview of notable fact-checking frameworks for LLMs and their approach to mitigate hallucination.}
\label{tab:frameworksforhallucinationmitigation}
\begin{tabularx}{\textwidth}{lX}
\toprule
\textbf{Framework} & \textbf{Hallucination Types Addressed} \\
\midrule
MiniCheck & Specifically targets factual errors where LLM-generated text is not grounded in or is inconsistent with provided source documents. It is designed to handle complex errors that require synthesizing information across multiple sentences and verifying multiple facts within a single claim. \\
\addlinespace
CliniFact & Addresses factual errors in the highly specialized medical domain by verifying clinical research claims. It also targets errors in Logical Reasoning, as it evaluates logical statements derived from hypothesis testing in clinical trials. \\
\addlinespace
VisualFactChecker & Explicitly designed to mitigate hallucinations in image captioning, such as generating descriptions of non-existent objects (content) or incorrect attributes (shape, color), by using visual tools (object detection, VQA) for verification. \\
\addlinespace
FACT-GPT & Addresses the challenge of recurring misinformation by matching new social media posts to previously debunked claims. It tackles variations of false claims that entail or contradict a known falsehood. \\
\addlinespace
Hierarchical Step-by-Step (HiSS) & This method uses a prompting technique to mitigate hallucination, where prompts are given to LLMs to do fine-grained checking of news claims. Here, prompts are given to the LLM to decompose the claim into subclaims and verify each subclaim step-by-step by raising and answering a series of questions. For each question, a prompt is again given to the LLM to assess if it is confident to answer it or not, and if not, the question is given to a web search engine. The search results are then inserted back into the ongoing prompt to continue the verification process. \\
\addlinespace
LLM-KG Framework & Mitigates hallucinations by grounding LLM responses in a structured and verifiable Knowledge Graph. It helps prevent the generation of plausible but incorrect facts by providing explicit, verified triplets as context. \\
\addlinespace
LLM-AUGMENTER & Mitigates hallucinations by augmenting a fixed LLM with external knowledge to generate grounded responses. It uses automated feedback to iteratively revise responses and improve their factuality score against the evidence. \\
\addlinespace
OpenFactCheck & Evaluates and identifies various types of factual errors, including Knowledge Errors (hallucinated/inaccurate information), Over-commitment Errors (failing to recognize false premises), and Disability Errors (outdated information). \\
\addlinespace
MEDICO & This is a framework specifically designed for hallucination detection and correction. It fuses evidence from multiple sources (web, knowledge bases, knowledge graphs) to detect factual errors, provide a rationale, and iteratively revise the hallucinated content. \\
\addlinespace
Yours Truly & This framework is designed to fact-check social media claims by breaking down compound sentences into atomic claims and verifying each against a real-time database of fact-checked articles. It addresses the challenge of hallucinations in LLMs by verifying outputs against an external, curated knowledge source (FactStore). \\
\addlinespace
Logical Consistency Framework & This framework addresses inconsistencies in LLM responses, which are a vulnerability related to hallucination. It assesses and improves the logical consistency of LLMs when performing fact-checking against Knowledge Graphs on complex queries involving logical operators, aiming to make the LLM less likely to hallucinate by ensuring its reasoning is logically sound. \\
\bottomrule
\end{tabularx}
\end{table*}

In terms of instruction-following techniques for factuality, several notable works, including Cheung et al. \cite{cheung2023factllama}, explicitly enhanced instruction-following models by integrating external knowledge specifically for fact-checking. They illustrated that targeted fine-tuning can significantly improve factual accuracy. Setty et al. \cite{setty2024surprising} demonstrated that smaller, finely-tuned models can sometimes surpass larger, more general models in accuracy for fact-checking tasks. Tran et al. \cite{tran2024leaf} introduced two parallel self-training methods to update LLMs and boost factual reliability: Supervised Fine-Tuning (SFT) using verified responses, and Simple Preference Optimization (SimPO) using fact-based ranking. Qi et al. \cite{qi2024sniffer} proposed a two-stage instruction tuning method to adapt InstructBLIP for OOC misinformation detection, first aligning it to the news domain using NewsCLIPpings data \cite{luo2021newsclippings} and then fine-tuning it on GPT-4-generated inconsistency explanations. Leite et al. \cite{leite2023detecting} introduced a multi-stage weak supervision approach utilizing instruction-tuned LLMs prompted with 18 credibility signals to generate weak labels, which are then aggregated to predict content veracity, thereby minimizing hallucinations and enhancing transparency for human fact checkers. Jing et al. \cite{jing2024scale} fine-tuned two HHEM DeBERTa NLI models on synthetic data and found that the HHEM model with fine-tuning on synthetic data can outperform LLMs in domain-specific evaluation, given carefully crafted training data.

\vspace{0.02\linewidth}\noindent\textbf{\textit{Retrieval-Augmented Generation.}} RAG has emerged as a prominent technique to ground LLM outputs in external, verifiable knowledge sources, with the aim of improving the factual context, reducing hallucination, and reliance on \tb{potentially flawed internal knowledge \cite{augenstein2024factuality, si-etal-2024-large, quelle2024perils, singhal2024evidence, tharaniya2025ontology}}. RAG architectures generally include a retriever module that gathers relevant information from external sources, such as web documents or databases, and a generator module (the LLM) that synthesizes this retrieved information and formulates answers or assessments \cite{peng2023check, sankararaman2024provenance}. 

Several studies have demonstrated the effectiveness and variations of RAG. For instance, Singhal et al. \cite{singhal2024evidence} reported how integrating external knowledge and feedback loops can significantly improve factuality. Table \ref{tab:quantitative_summary} illustrates these performance gains across multiple benchmarks, including PolitiFact and PubMedQA. Peng et al. \cite{peng2023check} introduced LLM-AUGMENTER that enhances LLM responses by retrieving external knowledge, linking raw evidence with related context, verifying outputs, and iteratively refining prompts using automated feedback until the response is free from hallucination and factually grounded. On the other hand, Quelle et al. \cite{quelle2024perils} allowed LLMs to perform Google searches to gather evidence related to the claim. Sankararaman et al. \cite{sankararaman2024provenance} presented a method, Provenance, where a dual-stage cross-encoder framework, one stage filters and weights context items, and the other assesses the factuality score, with the scores aggregated for threshold-independent evaluation. Khaliq et al. \cite{khaliq2024ragar} applied multimodal reasoning within RAG, proposing Chain of RAG (CoRAG) and Tree of RAG (ToRAG), specifically for political contexts.

Reinforcement learning has also been used in the RAG process. For example, Zhang et al. \cite{zhang2024reinforcement} investigated reinforcement learning to optimize retrieval processes. \cite{xie2024fire} proposed an iterative retrieval and verification mechanism to refine accuracy further. Singhal et al. \cite{singhal2024evidence} introduced a RAG-based fact-checking system where the core pipeline retrieves top-3 documents with FAISS, extracts evidence, and then uses the evidence in classification, combining the In-Context Learning (ICL) capabilities of multiple LLMs, achieving a 22\% gain on the Averitec dataset. Tran et al. \cite{tran2024leaf} introduced Fact-Check-Then-RAG, which improves the factual accuracy of LLM by evaluating each fact with SAFE \cite{Wei2024longform}, retrieving relevant data using ColBERT from MedRAG for failed checks, and regenerating responses via a RAG-enhanced prompt. Zhao et al. \cite{zhao2024pacar} introduced a multi-agent and retrieval-augmented fact-checking system that dynamically selects reasoning tools and external evidence to handle diverse multi-hop verification tasks. 

To improve out-of-context (OOC) detection, Qi et al. \cite{qi2024sniffer} integrate Google’s Entity Detection API for visual grounding and perform external verification through LLM to verify news captions against evidence retrieved from reverse image searches. Li et al. \cite{DBLP:conf/naacl/LiPGGZ24/self-checker} introduced SELF-CHECKER, which verifies input by extracting simple claims, generating search queries on various external knowledge sources, e.g., Bing Search API, retrieving evidence from knowledge sources, e.g., Wikipedia, Reddit messages, and predicting each claim's veracity based on its selected evidence sentences. Giarelis et al. \cite{giarelis2024unified} proposed a unified LLM-KG framework for fact-checking, which retrieves relevant facts from Knowledge Graphs (KGs) and injects them into the LLM prompt for response generation. Ma et al. \cite{ma2025local} introduced a Logical and Causal fact-checking method (LoCal), a LLM-driven multi-agent framework that breaks down complex claims, resolves them through specialized reasoning, and validates consistency using logical and counterfactual evaluators in an iterative process. Zhao et al. \cite{zhao2024medico} introduced MEDICO, where the system retrieves evidence from various sources, search engines, knowledge bases (KB), KGs, and user files, then re-ranks and fuses it by concatenation or Llama3-8B-based summarization. Li et al. \cite{li2024large} introduced FactAgent that uses LLM's internal knowledge (i.e., Phrase, Language, Commonsense, Standing tools) and another that integrates external knowledge tools (i.e., URL and Search tools - SerpAPI). Ghosh et al. \cite{ghosh2024logical} introduced LLMQuery that evaluates LLMs' logical consistency in fact-checking by retrieving subgraphs from KGs using BFS or ANN-based vector embedding methods, ensuring concise and relevant context is fed to the model. Finally, Tang et al. \cite{tang2024minicheck} introduce an efficient method specifically tailored to evaluate claims generated by LLM against grounding documents, which forms a core component of a comprehensive evaluation of RAG.

\vspace{0.02\linewidth}\noindent\textbf{\textit{Adversarial Tuning.}} Adversarial training presents LLMs with specifically designed examples intended to uncover hallucinations or factual errors, thus training the models to recognize and handle these challenging inputs accurately and improve their robustness against generating misinformation. The study by Leippold et al. \cite{leippold2025automated} introduced CLIMINATOR, an acronym for CLImate Mediator for INformed Analysis and Transparent Objective Reasoning, and an AI-based tool utilizing a Mediator and adversarial Advocate framework to automate climate claim verification by simulating structured debates, including climate denial perspectives, iteratively reconciling diverse viewpoints to converge towards scientific consensus consistently and thus improving accuracy and reliability.

\vspace{0.02\linewidth}\noindent\textbf{\textit{Automated Feedback Mechanisms and Self-Correction.}} Incorporating automated feedback loops and enabling LLMs to self-critique and correct their outputs are emerging as powerful strategies, exemplified by several key approaches. Peng et al. \cite{peng2023check} proposed an automated feedback mechanism, LLM-AUGMENTER, which substantially reduces ChatGPT's hallucinations without sacrificing the fluency and informativeness of its responses by iteratively revising LLM prompts to improve model responses using feedback generated by utility functions, e.g., the factuality score of a LLM-generated response. Ma et al. \cite{ma2025local} utilized two evaluating agents that iteratively reject or accept solutions and trigger new decomposition or reasoning rounds until consistency is achieved. Additionally, iterative refinement processes, such as those demonstrated by Xie et al. \cite{xie2024fire}, continuously check and improve model outputs through multiple verification cycles; Tran et al. \cite{tran2024leaf} introduced LEAF, a self-training loop that utilizes fact-check scores as automated feedback. 

Furthermore, the uncertainty quantification approach proposes using token-level uncertainty measures to detect potential hallucinations and trigger additional verification steps, thereby enhancing overall reliability and accuracy \cite{fadeeva2024fact}. Ge et al. \cite{ge2024visual} also showed in their framework that LLMs utilize object-detector and VQA results to mitigate hallucinated contents automatically and fact-check proposed captions. In the work of Zhao et al. \cite{zhao2024medico}, it iteratively corrects only hallucinated parts in generated content using Chain-of-Thought (CoT) prompting, while enforcing minimal edits via Levenshtein-based preservation scoring.

\vspace{0.02\linewidth}\noindent\textbf{\textit{Hybrid Approaches and Multi-Agent Systems.}} Combining multiple strategies or employing multi-agent architectures, wherein different LLM agents handle specialized sub-tasks within the fact-checking process, has emerged as a growing trend, exemplified by hierarchical prompting and planning methods such as the work by Zhang et al. \cite{zhang2023towards} systematically guided models through complex claim verification, and Zhao et al. \cite{zhao2024pacar} utilized LLMs for structured planning and reasoning tasks. 

Additionally, multi-agent systems have gained attention, as seen in studies like LoCal \cite{ma2025local}, employing multiple specialized LLM agents (decomposer + reasoner + two evaluators) to address logical and causal dimensions of fact-checking. Further explored by Kupershtein et al. \cite{kupershtein2025ai} and Li et al. \cite{li2024large}, both of which investigate the capabilities of agent-based frameworks specifically for fake news detection. Li et al. \cite{li2024large} also introduced FactAgent that behaves in an agentic manner, emulating human expert behavior utilizing several tools (Phrase, Language, Commonsense, Standing, URL, and Search tools) around a single LLM. 

Moreover, the integration of structured knowledge sources is further explored, advocating the combination of LLMs with knowledge graphs to comprise structured factual information \cite{giarelis2024unified}. Additionally, Hu et al. \cite{hu2024bad} designed a novel approach, ARG and its distilled version ARG-D, that complements small and large LMMs by selectively acquiring insights from LLM-generated rationales for SLMs. This combination of LLM+SLM has shown superiority over existing SLM/LLM-only methods. By introducing a hybrid KG retrieval + LLM generation approach and supervised fine-tuning, Ghosh et al. \cite{ghosh2024logical} improved the logical consistency of LLMs on the complex fact-checking task. Similarly, Jing et al. \cite{jing2024scale} combined rubric-prompted LLM judges and an NLI cross-encoder (HHEM) in the same framework.

\subsubsection{Recent Innovations for Reducing Hallucinations and Improving Factuality}

Beyond the above core strategies, recent studies have further tried to address hallucination issues and ensure LLMs remain faithful to facts. These methods range from innovative prompting techniques and multi-step reasoning procedures to incorporating multiple modalities, as well as building self-checking mechanisms and uncertainty estimates into LLM responses. We highlight several promising directions below, along with their advances in architecture, evaluation, and practical deployment.

\vspace{0.02\linewidth}\noindent\textbf{\textit{Multimodal Fact-Checking.}} Misinformation is increasingly multimodal, combining text, images, and videos, thus addressing hallucinations and ensuring factuality in LLMs processing such data has become crucial. Efforts to integrate visual and textual evidence include the study by Cao et al. \cite{cao2024multi}, which employed graph attention networks to consolidate multimodal knowledge for verifying claims, and Ge et al. \cite{ge2024visual}, focusing specifically on accurate captioning of images utilizing visual fact-checking through object detection and VQA models. Then, to address multimodal fact-checking with explanation generation, Yao et al. \cite{yao2023end} presented Mocheg, a large-scale benchmark dataset, and an initial demonstration of existing methods' performance, which generated unsatisfactory results. Subsequently, Papadopoulos et al. \cite{papadopoulos2025red} proposed the RED-DOT model, integrating a module called "relevant evidence detection" (RED) that can evaluate the relevancy of each piece of evidence and achieves up to 33.7\% accuracy gain on the VERITE benchmark. Furthermore, Sharma et al. \cite{sharma2024vision} evaluated the visual grounding capabilities inherent in language models, while \cite{qi2024sniffer} introduced a multimodal LLM-SNIFFER which analyzes both the consistency of the image-text content and the claim-evidence relevance using InstructBLIP and GPT-4V. Fake news detection is an emerging and vital domain addressed by Wang et al. \cite{wang2024llm}, where they proposed a hybrid model called FND-LLM that combines SLMs and LLMs. This results in significant accuracy improvements of 0.7\%, 6.8\%, and 1.3\% on the Weibo \cite{jin2017multimodal}, Gossipcop \cite{hu2024bad}, and Politifact datasets. Recently, Kakizaki et al. \cite{Kakizaki25maft} introduced a system called Multimodal Automated Fact-checking via Textualization (MAFT), that textualizes content such as images, video, and audio for LLM analysis, which generates comprehensive and interpretable reports that explain the verification steps to combat misinformation. Practical applications are explored in \cite{geng2024multimodal}, which investigates real-world deployment scenarios of multimodal LLMs, and finally, Khaliq et al. \cite{khaliq2024ragar} specifically applied retrieval-augmented reasoning (ToRAG, CoRAG) to tackle multimodal claims within political context by extracting both textual and image content, retrieving external information, and reasoning subsequent questions to be answered based on prior evidence and achieved a weighted F1-score of 0.85, surpassing a baseline reasoning method by 0.14 points.

\vspace{0.02\linewidth}\noindent\textbf{\textit{Multilingual Fact-Checking.}} Misinformation transcends language barriers, necessitating fact-checking capabilities across multiple languages. Zhang et al. \cite{zhang2023dont} in their work conducted a systematic quantitative and qualitative analysis of the multilingual abilities of LLMs by introducing a novel prompt back-translation method. Their result reveals that LLMs excel in transferring learned knowledge across different languages, producing relatively consistent results in translation-equivariant tasks but struggle with translation-variant tasks, requiring careful human evaluation. Therefore, assessing the cross-lingual factual consistency of LLMs should be a primary concern in current research. For instance, Shafayat et al. \cite{shafayat2024multifact} proposed a pipeline called Multi-FAct to evaluate the factuality of long-form multilingual LLM generations. This pipeline demonstrates the feasibility of adapting FActScore with non-English resources as a knowledge source, highlights their potential and limitations, and investigates cultural and geographic biases in LLMs. Quelle et al. \cite{quelle2024perils} found that fact-checking accuracy varied across languages, with translated English prompts often achieving higher accuracy than original non-English ones in terms of multilingual fact-checking, despite claims involving non-English sources. Additionally, due to skewed training data and non-standardized fact-checks, LLMs perform better with English-translated prompts, revealing language bias in multilingual fact verification. We also found a unique and noteworthy work by Siino et al. \cite{siino2022fake} which introduced a comparative evaluation of state-of-the-art models, including LLMs for detecting fake news. Utilizing the multilingual Profiling Fake News Spreaders on Twitter (PFNSoT) dataset, which includes English and Spanish tweets, they showed that large, pre-trained Transformer models such as RoBERTa, DistilBERT, BERT, XLNet, ELECTRA, and Longformer are not necessarily the optimal solution for the classification task; instead, a convolutional neural network (CNN) is enough to outperform those. Their claim was validated by their proposed CNN model, demonstrating a binary accuracy of 71.5\% for the English dataset and 81.5\% for the Spanish dataset, where the highest performing transformer models, RoBERTa, only achieved an accuracy of 69.5\% in English and ELECTRA achieved 76\% in Spanish. Notably, in recent years, Shcharbakova et al. \cite{shcharbakova2025scale} demonstrated a comprehensive benchmark of language models for multilingual fact verification. Their experiments showed that the smaller encoder-based language model XLM-R significantly outperformed a much larger decoder-only LLM, achieving a macro-F1 score of 57.7\%, which represents an improvement of approximately 15.8\% over the previous state-of-the-art results. They suggested specialized SLMs can be more effective than general-purpose LLMs but also mentioned systematic difficulties in utilizing evidence effectively, alongside notable biases toward frequent categories within imbalanced data settings. In multilingual fact-checking, we found a distinctive and noteworthy work, performed by Jannah et al. \cite{jannah2025multilingual}, where they assessed the symptom detection capabilities of LLMs in social media posts across seven languages. Their experiment on zero-shot multi-label symptom classification includes large- and small-parameter models from three leading LLM providers: OpenAI, Google Gemini, and Mistral AI. They showed significant performance disparities, with LLMs performing better in European languages than in under-resourced Asian languages, and tending to over-predict specific respiratory illnesses like influenza.

\vspace{0.02\linewidth}\noindent\textbf{\textit{Domain-Specific Fact-Checking.}} Domain-specific fact-checking is a crucial research area, as the nuances of verifying factual claims can significantly differ across specialized fields like medicine, politics, and climate science, necessitating tailored LLMs and verification systems. While general-purpose fine-tuned LLMs dominate broad tasks, specialized models fine-tuned on specific domains often outperform general-purpose models in those areas \cite{zhang2025dataset}. Moreover, proprietary models like Factcheck-GPT are often designed for general-purpose use and not viable in domains like medicine due to restrictions on private data use and lack of fine-tuning \cite{tran2024leaf}.

In medical contexts, dedicated efforts include \cite{zhang2025dataset}, introduced CliniFact and when evaluating LLMs against that, BioBERT achieved 80.2\% accuracy, outperforming generative counterparts, such as Llama3-70B’s 53.6\%, with statistical significance (p < 0.001), developing explainable reasoning systems \cite{vladika2025step}, and detecting and correcting hallucinations by integrating multi-source evidence \cite{zhao2024medico}. Similarly, Tans et al. \cite{tran2024leaf} proposed LEAF, which is tailored towards the medical domain as it uses the MedRAG corpus. 

Fact-checking is also investigated across various domains, including travel, climate, news information, and claim matching. For example, in the travel domain, Jing et al. \cite{jing2024scale} used four industry datasets containing chats, reviews, and property information for fact-checking. Political claim verification is explored through studies assessing LLMs' reliability \cite{chatrath2024fact} and multimodal retrieval-augmented reasoning systems \cite{khaliq2024ragar}. In climate science, specialized LLM-based tools address the complexity of climate-related claims \cite{choi2024automated}. News claims and general misinformation are broadly examined through hierarchical prompting methods in \cite{zhang2023towards}, \cite{hu2024bad}. Other studies investigated the potential of LLMs for fake news detection using datasets such as Weibo21 \cite{nan2021mdfend} and GossipCop \cite{Shu2018FakeNewsNet}. Research on news-article veracity has focused on the FA-KES Syrian War corpus \cite{salem2019fa} and the EUvsDisinfo pro-Kremlin corpus \cite{EUvsDisinfo}, while some platforms enable the development of customized fact-checking systems \cite{wang2025openfactcheck}. \cite{qi2024sniffer} trained their SNIFFER model on news domain using the NewsCLIPpings \cite{luo2021newsclippings} dataset. Additionally, claim-matching methods utilizing LLMs for fact-checking \cite{pisarevskaya2025zero, choi2024automated, choi2024fact} and verification approaches for complex claims \cite{liu2025bidev} further underscore the depth and breadth of ongoing domain-specific fact-checking research.

\vspace{0.02\linewidth}\noindent\textbf{\textit{Enhancing Explainability and Trust.}} Beyond mere accuracy, the ability of LLM-based fact-checking systems to provide explanations and foster trust is increasingly recognized as vital. Citations and provenance play a central role in this, with \cite{ding2025citations} highlighting the importance of referencing sources to build user confidence, while \cite{sankararaman2024provenance} emphasizes the need to trace the origin of generated content. In the framework proposed by Quelle et al. \cite{quelle2024perils}, agents explain their reasoning and cite the relevant sources. Zhao et al. \cite{zhao2024pacar} enhanced fact-checking explainability by utilizing instruction-based LLMs and specialized agents to generate structured reasoning and justifications for sub-claims. Through CLIP-based similarity, ROUGE scores, response ratio analysis, and human evaluation, SNIFFER \cite{qi2024sniffer} demonstrated high accuracy and strong persuasive ability in explaining and detecting out-of-context misinformation, and transparency in medical contexts \cite{vladika2025step}. Credibility assessment frameworks offer structured approaches to evaluate trustworthiness \cite{krishnamurthy2024yours}. Additionally, Ghosh et al. \cite{ghosh2024logical} investigated the logical soundness of model outputs, a key indicator of reliability. In another study, Leite et al. \cite{leite2023detecting} simplified fact-checking by guiding LLMs to predict individual veracity signals, minimizing hallucinations, and enabling human reviewers to audit and control outputs, which enhanced transparency. Giarelis et al. \cite{giarelis2024unified} stated that their LLM-KG framework improves transparency through factual context provided by KGs.

\vspace{0.02\linewidth}\noindent\textbf{\textit{Hierarchical Prompting and Multi-Step Reasoning.}} A key innovation involves prompting LLMs in a way that structures their reasoning process by decomposing complex fact-checking tasks into smaller, verifiable steps. The underlying idea is that human fact-checkers often break down a claim into sub-claims or evidence checks. LLMs can be prompted to emulate the process, reducing the risk of a single misstep leading to a hallucinated conclusion. For instance, HiSS prompting directs the model to first separate a claim into several subclaims, then verify each subclaim one by one, before finalizing an overall verdict \cite{zhang2023towards}. By forcing the model to focus on one piece of information at a time (in a chain-of-thought style), HiSS achieved superior fact verification performance and reduced hallucination on news datasets, even outperforming fully-supervised baselines. Advance prompting like Hiss outperforms standard CoT Table \ref{tab:quantitative_summary}. This demonstrates that well-designed prompting can guide the model to reason more carefully and factually reduce hallucination. In a study, Khaliq et al. \cite{khaliq2024ragar} introduced Chain-of-RAG (sequential) and Tree-of-RAG (branch-and-eliminate hierarchy), which embodied multi-step and hierarchical reasoning. Another example is the PACAR framework \cite{zhao2024pacar}, which combined LLM-driven planning with customized action reasoning for claims. PACAR consists of multiple modules (a claim decomposer, a planner, an executor, and a verifier) that allow LLMs to plan a sequence of actions, such as performing a web search or a numerical calculation, and then verify the claim based on collected evidence. Using hierarchical prompting and a multi-step approach, which includes specialized skills like numerical reasoning and entity disambiguation, PACAR significantly outperformed baseline fact-checkers across three different domain datasets. 

Overall, research to date illustrates the use of several comprehensive, multi-dimensional approaches to mitigating LLMs' hallucinations in fact-checking, with notable advances in RAG domain-specific fine-tuning and hybrid methodologies. Nonetheless, guaranteeing robust factual reliability across varied, complex, and dynamically evolving information scenarios continues to pose a major challenge. Future research is expected to prioritize more sophisticated hybrid systems, refined self-correction mechanisms, and more effective human-AI collaboration to strengthen the fact-checking processes \cite{si-etal-2024-large, yang2024fact}.

\subsection{Datasets for Training and Evaluating Fact-Checking Systems (RQ3)}
\label{dataset_section}

In this section, we discuss the wide range of datasets used in the training, evaluation, and benchmarking of fact-checking systems, particularly within RAG frameworks and hallucination mitigation strategies. These datasets support various steps in each method, such as claim verification, evidence retrieval, multi-hop reasoning, and hallucination detection. The following accounts for the datasets and their uses in this domain:
% [repeated info in the bulletpoints -- removing it]
% Below, we have categorized the datasets by specific methods of fact-checking.

% \begin{itemize}
%     \item \textbf{Claim-Evidence Pair Datasets}: e.g., FEVER, SciFact
%     \item \textbf{Multi-hop or Structured Reasoning Benchmarks}: e.g., HOVER, FEVEROUS
%     \item \textbf{Domain-Specific and Biomedical QA Sets}: e.g., MedMCQA, BioASQ
%     \item \textbf{Multimodal Fact-Checking Datasets}: e.g., MM-FEVER, Post-4V
%     \item \textbf{Synthetic and Weakly Supervised Corpora}: e.g., ClaimMatch, LLM-AGGREFACT
%     \item \textbf{Composite Benchmarks}: e.g., FactBench, FIRE
%     \item \textbf{Multilingual or Low-Resource Collections}: e.g., X-Fact, FactStore
% \end{itemize}
\vspace{0.02\linewidth}\noindent\textbf{\textit{Benchmark Datasets for RAG-Based Fact Verification.}} Core claim verification datasets such as FEVER \cite{thorne2018fever}, FEVEROUS \cite{aly2021feverous}, and HOVER \cite{jiang2020hover} are widely used to evaluate RAG pipelines, where the model is used to retrieve relevant evidence from Wikipedia or structured sources and then generate a verdict. These datasets provide gold-standard evidence, making them ideal for training retrievers and verifying the accuracy of generated outputs. LIAR \cite{wang2017liar} and RAWFC \cite{zhang2022cofced} further allow the assessment of RAG-based models on political and news-based claims with distinctive complexity and source structures.

\vspace{0.02\linewidth}\noindent\textbf{\textit{Domain-specific Datasets.}} In domain-specific applications, datasets such as SciFact \cite{wadden2020fact}, COVID-Fact \cite{saakyan2021covid}, MedMCQA \cite{pal2022medmcqa}, BioASQ \cite{tsatsaronis2015overview}, and PubMedQA \cite{jin2019pubmedqa} are frequently employed in RAG frameworks that are aligned with the biomedical domain. These allow models to retrieve evidence from the medical literature (e.g., via MedRAG) and cross-validate LLM outputs. Given the importance of factual accuracy in these domains, these datasets also serve as valuable benchmarks for evaluating and refining hallucination reduction techniques in sensitive contexts.

\vspace{0.02\linewidth}\noindent\textbf{\textit{Multimodal Datasets.}} Multimodal datasets such as Multimodal-FEVER (Factify) \cite{mishra2022factify}, Post-4V \cite{geng2024multimodal}, NewsCLIPpings \cite{luo2021newsclippings}, and MOCHEG \cite{yao2023mocheg} allow extended fact-checking to vision-language settings. These are especially relevant for evaluating multimodal RAG systems that allow the use of visual and textual information to assess the veracity of claims. Mismatches between image-caption pairs in these datasets test the models’ ability to detect hallucinated or manipulated content across modalities.

\vspace{0.02\linewidth}\noindent\textbf{\textit{Hallucination Detection-specific datasets.}} To evaluate hallucination detection and correction, specialized datasets such as HaluEval \cite{li2023halueval}, ReaLMistake \cite{kamoi2024realmistake}, TruthfulQA \cite{lin2021truthfulqa}, and FoolMeTwice \cite{eisenschlos2021foolmetwice} provide annotated examples of hallucinated outputs with detailed rationales. These are critical for assessing token-level uncertainty, logical consistency, and explanation alignment in LLMs.

\vspace{0.02\linewidth}\noindent\textbf{\textit{Composite Datasets.}}
Composite benchmarks like FactBench \cite{zhang2024factbench}, OpenFactCheck \cite{wang2025openfactcheck}, and FIRE \cite{zhang2024fire} aggregate multiple datasets (e.g., FacTool-QA, FELM-WK, Factcheck-Bench) to provide diverse evaluation techniques for both retrieval and generation stages. These are specifically valuable for end-to-end RAG evaluation, as they test factuality, consistency, and explainability across all claim types and evidence formats.

\begin{figure}[!ht]
    \centering
    \includegraphics[width=\linewidth]{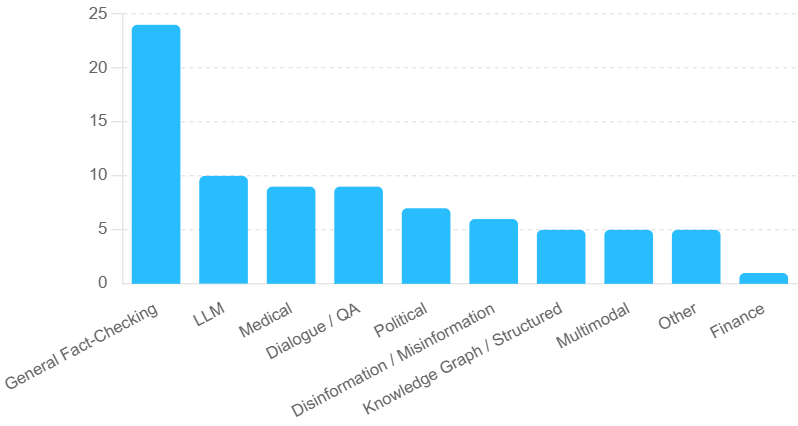}
    \caption{Illustration of major dataset types and domains. The bar chart shows the distribution of datasets across different types and application domains, highlighting their relative prevalence and focus areas.}
    \label{fig:dataset_types}
\end{figure}

\vspace{0.02\linewidth}\noindent\textbf{\textit{Synthetic and Multilingual Datasets.}} Synthetic and weak supervision datasets such as ClaimMatch \cite{panchendrarajan2025multiclaimnet}, LLM-AGGREFACT \cite{tang2024minicheck}, and CheckThat \cite{checkthat2022} allow for scalable training and evaluation in low-resource settings. These datasets are often used to pre-train or fine-tune retrievers and scorers within RAG systems, or to assess robustness against adversarial claims and misinformation edits. Additionally, multilingual datasets and some databases, like X-Fact \cite{gupta2021xfact}, Data Commons Multilingual, and FactStore, help build fact-checking systems that work across different languages. They test whether these systems can find the correct information and give accurate answers, even in non-English settings. This helps make fact-checking tools more significant in global use. Figure \ref{fig:dataset_types} visualizes an overview of major dataset types and their domains.

% [AR: 72 key datasets; RAG: 63; Halluc.: 48]
% % To provide a clear and comprehensive overview, Table \ref{tab:dataset-summary-split} lists 74 key datasets used in fact-checking research. It shows which datasets were used for fact checking by using RAG, for hallucination reduction, and their type. Out of 74 datasets, 59 use RAG and 52 are used for hallucination reduction. 
To provide a clear and comprehensive overview, Table \ref{tab:dataset-summary-split} lists 72 key datasets used in fact-checking research. It shows which datasets were used for fact-checking by using RAG, for hallucination reduction, and their type. Out of 72 datasets, 63 use RAG, and 48 are used for hallucination reduction. The use cases of each dataset used for fact-checking are grouped into several key types. To test interactive evidence retrieval and conversational-based fact-checking in multi-turn RAG pipelines, dialogue datasets are used \cite{chatrath2024fact,ji2023survey}. QA datasets test a system's ability to accurately answer questions by reasoning through open-domain and multi-hop scenarios, where it retrieves and connects relevant information \cite{huang2021factual,jiang2020hover}. Fact-checking datasets provide labelled claims with the most reliable evidence for training and benchmarking claim verification \cite{aly2021feverous,panchendrarajan2025multiclaimnet}. Complex fact-checking in biomedical contexts is typically facilitated by medical and domain-specific datasets (e.g., SciFact, PubMedQA) \cite{ji2023survey,jin2019pubmedqa}. LLM and hallucination datasets capture ungrounded outputs to assess truthfulness and hallucination mitigation. Structured data-to-text datasets can produce more accurate texts from tables or knowledge graphs \cite{ladhak2023pre}. Summarisation datasets evaluate the production of concise and reliable text \cite{augenstein2024factuality}. Misinformation datasets help to find and disprove false or manipulated claims \cite{Shu2018FakeNewsNet,checkthat2022}, and multimodal datasets check that text–image evidence pairs are consistent \cite{ge2024visual,jin2017multimodal}.

\begin{table*}[ht!]
\centering
\begin{scriptsize}
\caption{Datasets and their use in RAG and hallucinations (Halluc.) reduction. It indicates the type of dataset and the tasks for which it has been evaluated.}
\label{tab:dataset-summary-split}
\begin{tabular}{p{3cm}>{\centering\arraybackslash}p{0.8cm}>{\centering\arraybackslash}p{1cm}p{2.2cm}|p{3cm}>{\centering\arraybackslash}p{0.8cm}>{\centering\arraybackslash}p{1cm}p{2cm}}
\midrule
\textbf{Dataset} & \textbf{RAG} & \textbf{Halluc.} & \textbf{Type} & 
\textbf{Dataset} & \textbf{RAG} & \textbf{Halluc.} & \textbf{Type} \\
\midrule
Doc2Dial          & \cmark & \cmark & Dialogue & LIAR              & \cmark & \xmark  & Political \\
Topical-Chat   & \cmark & \cmark & Dialogue & RAWFC             & \cmark & \xmark  & Political \\
QReCC             & \cmark & \cmark & Dialogue & HALLU          & \cmark & \cmark & QA \\
Wizard of Wikipedia & \cmark & \cmark & Dialogue & BioASQ-Factoid    & \cmark & \cmark & QA \\
CMU-DoG           & \cmark & \cmark & Dialogue & Q2             & \cmark & \cmark & QA \\
FEVER             & \cmark & \xmark  & Fact-Checking & ASSERT         & \cmark & \cmark & QA \\
SciFact           & \cmark & \xmark  & Fact-Checking & SQuAD          & \cmark & \cmark & QA \\
COVID-Fact        & \cmark & \xmark  & Fact-Checking & CoCoGen        & \xmark  & \cmark & QA \\
Factify           & \cmark & \xmark  & Fact-Checking & TriviaQA       & \cmark & \cmark & QA \\
AVERITEC          & \cmark & \xmark  & Fact-Checking & HotpotQA       & \cmark & \cmark & QA \\
TruthBench     & \cmark & \cmark & Fact-Checking & Natural Questions & \cmark & \cmark & QA \\
LLM-AGGREFACT     & \cmark & \xmark  & LLM & ELI5           & \cmark & \cmark & QA \\
TruthfulQA     & \cmark & \cmark & LLM & NarrativeQA    & \cmark & \cmark & QA \\
HaluEval       & \cmark & \cmark & LLM & NewsQA         & \cmark & \cmark & QA \\
BioASQ-Y/N        & \cmark & \xmark  & Medical & DROP           & \cmark & \cmark & QA \\
MedMCQA           & \cmark & \xmark  & Medical & FactMix           & \cmark & \cmark & QA \\
USMLE             & \cmark & \xmark  & Medical & DuoRC          & \cmark & \cmark & QA \\
MMLU-Medical      & \cmark & \xmark  & Medical & QuAC              & \cmark & \cmark & QA \\
PubMedQA          & \cmark & \xmark  & Medical & FactScore Dataset & \cmark & \cmark & QA \\
CliniFact         & \xmark  & \xmark  & Medical & Data Commons      & \cmark & \xmark  & Structured \\ 
MedQuAD           & \cmark & \cmark & Medical QA & WikiBio        & \cmark & \cmark & Structured Data \\ 
MedInfo QA        & \cmark & \cmark & Medical QA & ToTTo          & \cmark & \cmark & Structured Data \\
LiveQA-Medical    & \cmark & \cmark & Medical QA & WebNLG         & \cmark & \cmark & Structured Data \\
MEDIQA-RQE        & \cmark & \cmark & Medical QA & DART           & \cmark & \cmark & Structured Data \\
MeQSum            & \cmark & \cmark & Medical Summary & LogicNLG       & \cmark & \cmark & Structured Data \\
FakeCovid         & \xmark  & \xmark  & Misinformation & E2E NLG        & \cmark & \cmark & Structured Data \\ 
Multimodal FEVER  & \cmark & \xmark  & Multimodal & SAMSum         & \cmark & \cmark & Summarization \\
COCO              & \xmark  & \xmark  & Multimodal & FIED           & \cmark & \cmark & Summarization \\ 
Objaverse         & \xmark  & \xmark  & Multimodal & XSum           & \cmark & \cmark & Summarization \\
Visual Aptitude   & \xmark  & \xmark  & Multimodal & CNN/Daily Mail & \cmark & \cmark & Summarization \\
ADE20K            & \xmark  & \xmark  & Multimodal & Gigaword       & \cmark & \cmark & Summarization \\
NewsCLIPpings     & \cmark & \xmark  & Multimodal & Multi-News     & \cmark & \cmark & Summarization \\
Polyjuice      & \xmark  & \cmark & NLI & Newsroom       & \cmark & \cmark & Summarization \\
FactualNLI     & \cmark & \cmark & NLI & BigPatent      & \cmark & \cmark & Summarization \\
Multilingual FC   & \xmark  & \xmark  & Other & WikiHow        & \cmark & \cmark & Summarization \\
PolitiFact        & \cmark & \xmark  & Political & Reddit TIFU    & \cmark & \cmark & Summarization \\
\bottomrule
\end{tabular}
\end{scriptsize}
\end{table*}

\begin{table*}[ht!]
\centering
\caption{Identified Limitations of Datasets Commonly Used in LLM-Based Fact-Checking Across Different Domains}
\label{tab:dataset_limitations}
\begin{scriptsize}
\begin{tabularx}{\textwidth}{lX}
\hline
\textbf{Dataset Type} & \textbf{Key Limitations} \\
\hline
Dialogue & Low domain diversity and simple claim structures with weak coverage of real misinformation. \\

Fact-Checking & Biased toward Wikipedia claims with poor temporal and geographic variety and slow tracking of emerging misinformation. \\

QA & Encourages benchmark overfitting and lacks adversarial and multi-hop claims. \\

Medical & Expensive annotation with narrow coverage and poor multilingual and low-resource support. \\

Misinformation / Political & Small size and region specific, with rare updates for new narratives. \\

Structured Data & Clean and synthetic, but fails to capture noisy and conflicting real-world data. \\

Multimodal & Sparse and noisy, with weak standards and limited cross-modal reasoning tests. \\

LLM & Mostly synthetic with short and simple prompts and weak domain-specific coverage. \\

Summarization & Subjective references that reward lexical overlap over factual grounding. \\

Multilingual & Uneven language coverage with high-resource bias and weak cultural generalization. \\
\hline
\end{tabularx}
\end{scriptsize}
\end{table*}

Table \ref{tab:dataset_limitations} represents the key limitations of each dataset type used in fact-checking with LLM. Most datasets are narrow in scope or biased, which affects the reliability of fact-checking in specific domains and languages.

\subsection{Prompt Design, Fine-Tuning, and Domain-Specific Training (RQ4)}

\begin{figure*}[ht!]
    \centering
    \includegraphics[scale=0.13]{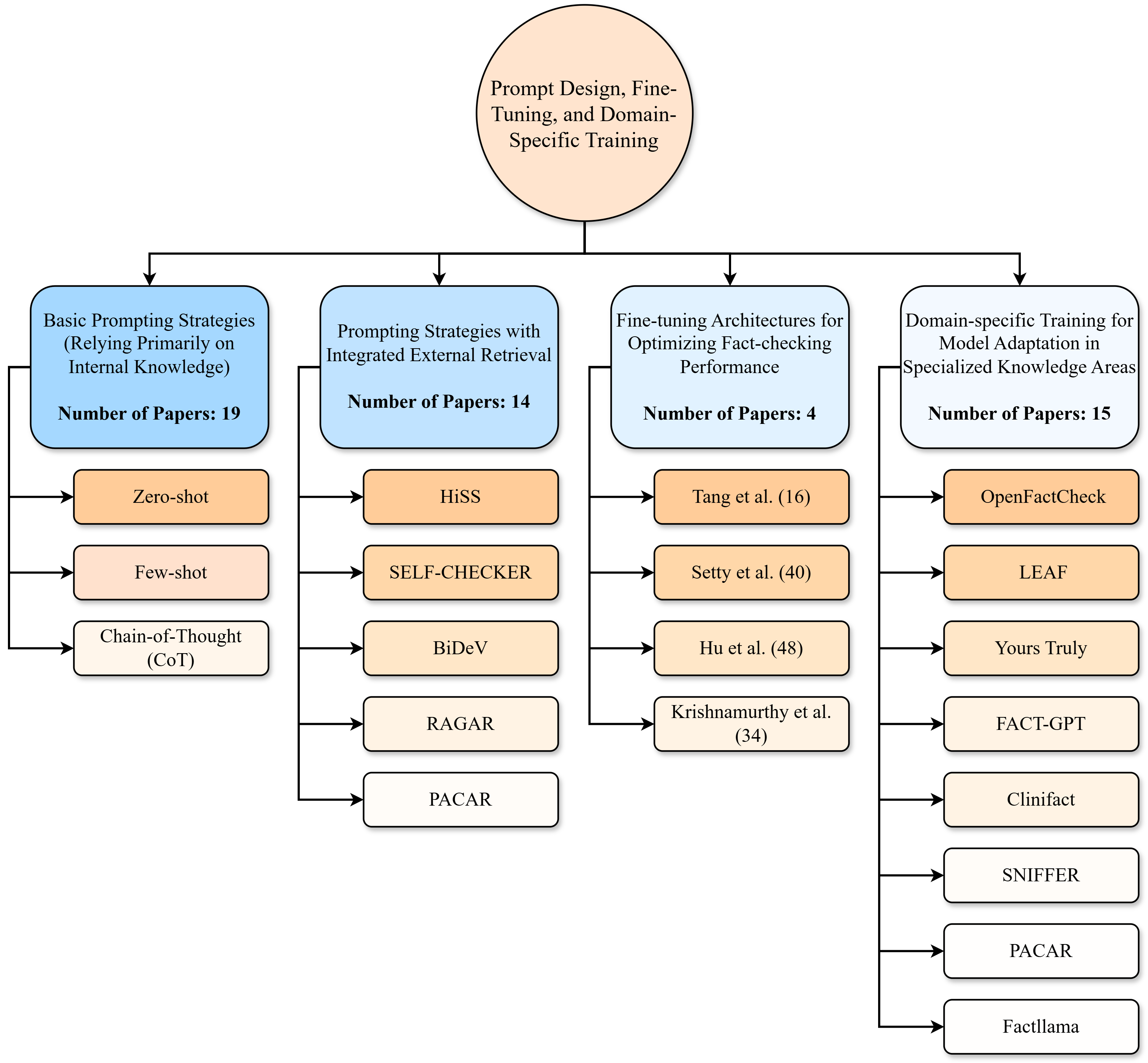}
    \caption{Breakdown of approaches in prompt design, fine-tuning, and domain-specific training for fact-checking with LLMs, categorized into four groups. Each group highlights the number of papers reviewed and the representative techniques.}
    \label{fig:section - 4.4}
\end{figure*}

Prompt design strategies significantly impact the ability of LLMs to perform fact-checking and mitigate hallucination. The choice of strategy influences how the model processes information, accesses knowledge, and generates responses, directly affecting accuracy and performance. The sources explore several key strategies, often contrasting methods that rely solely on the model's internal knowledge with those that integrate external information retrieval. A visual summary of approaches in prompt design, fine-tuning, and domain-specific training is shown in Figure \ref{fig:section - 4.4}.

\subsubsection{Basic Prompting Strategies}
Basic prompt strategies primarily rely on internal knowledge and involve presenting the claim or task to LLMs with minimal or no external context beyond the prompt itself. Their effectiveness is heavily dependent on the pre-trained knowledge of the model, which can be a significant limitation due to the potential for hallucination and outdated information \cite{ hu2024bad, zhang2023towards,tran2024leaf}.

Zero-shot prompting involves providing the LLMs with only the task description and the input claim, without any specific examples. This can include asking the model to directly predict the veracity label of a claim \cite{hu2024bad, li2024large}. Frameworks such as FACT-AUDIT \cite{lin2025fact} used zero-shot inference to evaluate the fact-checking capacity of various LLMs. They gave the lowest average accuracy scores compared to other methods, particularly when relying solely on the model's internal knowledge. While a simple combination of self-consistency and zero-shot prompt was found to be the most effective overall strategy in a multilingual fact-checking study, this effectiveness was strongly tied to the self-consistency decoding strategy, not necessarily the zero-shot nature itself \cite{singhal2024multilingual}. On the other hand, frameworks such as PACAR \cite{zhao2024pacar}, which employed explicit claim decomposition and a dynamic planning mechanism, achieved strong performance in zero-shot settings and outperformed other LLM-based approaches, including few-shot and fine-tuned methods. However, relying solely on internal knowledge for fact-checking is considered unreliable and insufficient, as LLMs are prone to hallucination \cite{ hu2024bad, zhang2023towards,tran2024leaf}. Zero-shot prompting without external access means that the model must rely on potentially inaccurate or outdated information stored during training \cite{quelle2024perils}. The analysis of rationales generated through zero-shot CoT indicates unreliability for factuality analysis based on internal memorization \cite{lee2020language}. The most significant mitigation discussed is the incorporation of external knowledge retrieval \cite{peng2023check, DBLP:conf/naacl/LiPGGZ24/self-checker, zhang2023towards, tran2024leaf, quelle2024perils,singhal2024evidence}.

The Few-Shot Prompting or ICL method involves providing the model with a limited number of examples of the task before presenting the claim to be verified \cite{hu2024bad, zhao2024pacar}. It utilizes the LLM's ability to learn from examples provided directly in the prompt ("in-context learning") \cite{DBLP:conf/naacl/LiPGGZ24/self-checker}. Few-shot demonstrations are used in methods like HiSS \cite{zhang2023towards} and BiDeV \cite{liu2025bidev} to guide the LLM through multi-step processes. Few-shot-CoT includes example pairs to guide the reasoning process \cite{choi2024automated, li2024large}. A similar approach was used to address the challenge of identifying hallucinations in natural language generation by using the pre-trained Llama 2-7B model in a zero-shot setting. Instead of fine-tuning the model, the authors used specific prompts that included the context and hypothesis sentences, which were provided to the model, to determine whether the context supported the hypothesis with a simple yes or no. The model's answers were reprocessed to classify them and determine hallucinations or unsupported outputs. They evaluated its performance on two tracks: the model-aware method, with access to specific model checkpoints, and the model-agnostic method, without such access. To keep the process efficient, the method focused on minimal engineering and prevented additional fine-tuning or preprocessing. They claim that the model's ability to reliably identify hallucinations could be improved even further with adjustments such as fine-tuning it for a specific domain or incorporating more data \cite{siino2024brainllama}. In another instance, an equivalent method is used in a prompt-based approach to few-shot learning to identify hallucinations in Swedish and English paraphrased texts using the Mistral 7B LLM. Task-specific prompts are carefully designed to include instructions and a limited number of reference examples from the training data. The model is given these prompts, the source sentence, and two hypotheses. Then, it is asked to determine which hypothesis contains hallucinated content. Without any additional fine-tuning, the method makes effective use of the model's reasoning abilities and strategically incorporates representative samples into the prompt to help the model make decisions that will successfully detect hallucinations \cite{siino2024gpt}.

ICL has been demonstrated to enhance the performance of open-source multimodal LLMs (MLLMs) in misinformation detection, occasionally yielding greater improvements than those achieved through prompt ensemble methods \cite{geng2024multimodal}. Combining ICL with RAG has been shown to improve accuracy in fact verification \cite{singhal2024evidence}. However, some sophisticated few-shot ICL methods like Standard Prompting, Vanilla CoT, and ReAct were surpassed by the HiSS method in news claim verification, highlighting the importance of the specific method prompted \cite{quelle2024perils}. While ICL improves performance, open-source MLLMs using ICL still significantly lag behind state-of-the-art proprietary models like GPT-4V. The effectiveness can vary between models and datasets \cite{geng2024multimodal}. The manual prompt design for few-shot examples can be heuristic \cite{DBLP:conf/naacl/LiPGGZ24/self-checker}. However, providing examples does not guarantee overcoming the dependence on internal knowledge if external information is not integrated \cite{zhang2023towards}. Integrating ICL with RAG or structured step-by-step prompting frameworks is a key mitigation \cite{ zhang2023towards,singhal2024evidence}.

Other methods, such as CoT prompting, instruct the LLM to output a sequence of intermediate reasoning steps before arriving at the final answer or the veracity label \cite{hu2024bad,zhang2023towards}. Zero-shot CoT prompts often include eliciting sentences like "Let us think step by step" \cite{hu2024bad,ma2025local, singhal2024multilingual, choi2024automated}. This encourages explicit reasoning. Variants of CoT prompting include English CoT (EN-CoT) \cite{singhal2024multilingual}, which focuses on monolingual reasoning, and CoTVP (CoT Veracity Prediction) \cite{khaliq2024ragar}, which evaluates the truthfulness of reasoning steps. However, studies show that techniques like CoT, designed to improve reasoning, do not necessarily improve the fact-checking abilities of LLMs. They can even have minimal or negative effects on success rates \cite{singhal2024multilingual}. Models prompted with CoT may tend to align with the input text rather than verifying its factualness, especially for complex paragraphs, when relying solely on pre-trained knowledge \cite{DBLP:conf/naacl/LiPGGZ24/self-checker}. Vanilla CoT suffers substantially from the issues of fact hallucination and omission of necessary thoughts in the reasoning process \cite{zhang2023towards}. Using the LLM for factuality analysis based on its internal memorization, even with zero-shot CoT, indicates unreliability, likely caused by hallucination \cite{hu2024bad}. 

The internal mechanism of LLMs to integrate rationales from various perspectives via CoT can be ineffective for fake news detection \cite{hu2024bad}. Integrating CoT with external knowledge retrieval, as in Search-Augmented CoT or ReAct \cite{zhang2023towards}, is a key approach to mitigate hallucination and thought omission \cite{zhang2023towards}. Furthermore, frameworks that explicitly guide the decomposition and reasoning process, such as FactAgent, are seen as superior to CoT, which primarily acts as a prompting technique \cite{li2024large}.

When relying solely on internal knowledge, LLMs prompted with zero-shot, few-shot, or vanilla CoT struggle in fact-checking complex claims and exhibit hallucination \cite{ hu2024bad, DBLP:conf/naacl/LiPGGZ24/self-checker, zhang2023towards,tran2024leaf}. However, accuracy can be low, and improvements in reasoning via CoT alone do not guarantee better fact-checking performance \cite{singhal2024multilingual}. Techniques for forcing binary "true" or "false" judgments also do not enhance overall accuracy \cite{deverna2024fact}.

\subsubsection{Prompting Strategies with Integrated External Retrieval}
Several strategies combine prompting with the ability to access and utilize external information sources (e.g., search engines or curated databases) to ground responses and improve factual accuracy. This is a critical aspect for robust fact-checking and hallucination reduction \cite{ zhang2023towards, tran2024leaf,quelle2024perils, singhal2024evidence}.

A variant of CoT that interleaves reasoning traces with task‑specific actions, like querying Google Search or the Wikipedia API, allowing the LLM agent to decide whether to search or continue reasoning based on environmental observations \cite{ DBLP:conf/naacl/LiPGGZ24/self-checker,zhang2023towards,khaliq2024ragar, quelle2024perils}. For instance, by accessing external knowledge, ReAct effectively mitigates hallucination failures compared to vanilla CoT and justifies its reasoning with retrieved citations, enhancing verifiability and explainability. Its performance is highly sensitive to the quality and relevance of search results, and relying solely on internal knowledge when external search fails remains a key limitation \cite{zhang2023towards}. Combining ReAct’s action capabilities with more structured decomposition and step‑by‑step verification methods, such as HiSS or SELF‑CHECKER, can address thought omissions and improve overall performance \cite{DBLP:conf/naacl/LiPGGZ24/self-checker, zhang2023towards}.

Search‑Augmented CoT augments vanilla CoT by using the original claim as a search query to retrieve background information, which the LLM incorporates into its thought chain. This approach improves over vanilla CoT by utilizing external knowledge, but can fall short of methods like Standard Prompting or HiSS, which indicates that querying solely with the claim may output insufficiently detailed results. To mitigate this, more sophisticated query generation strategies and integration methods are needed \cite{zhang2023towards}.
HiSS is a few‑shot method that prompts the LLM to perform claim verification in fine‑grained steps by decomposing claims into subclaims and verifying each step‑by‑step, raising questions and optionally using web search when confidence is low. It significantly surpasses few‑shot ICL counterparts like Standard Prompting, Vanilla CoT, and ReAct in average F1-score, offering superior explainability through enhanced coverage and readability while substantially reducing hallucination and thought omission \cite{zhang2023towards}. However, HiSS still struggles with integrating updated information from mixed sources and incurs high computational costs due to multiple LLM calls, with its performance remaining sensitive to prompt design \cite{DBLP:conf/naacl/LiPGGZ24/self-checker}.

Frameworks like SELF‑CHECKER \cite{DBLP:conf/naacl/LiPGGZ24/self-checker}, BiDeV \cite{liu2025bidev}, RAGAR \cite{khaliq2024ragar}, and PACAR \cite{zhao2024pacar} integrate prompting and RAG by decomposing fact‑checking into subtasks (e.g., claim detection, retrieval, sentence selection, verdict prediction) and using prompts often with few‑shot examples to generate search queries, select evidence, and perform step‑by‑step verification while explicitly incorporating retrieved documents. Incorporating external knowledge via RAG significantly boosts accuracy over internal‑only approaches: SELF‑CHECKER \cite{DBLP:conf/naacl/LiPGGZ24/self-checker}, BiDeV \cite{liu2025bidev}, RAGAR \cite{khaliq2024ragar}, and PACAR \cite{zhao2024pacar}  all show substantial gains, with specialized modules such as the Claim Atomizer and Fact‑Check‑Then‑RAG further enhancing predictive power and explanation quality. 

RAG‑based methods can be hampered by overwhelming context windows \cite{quelle2024perils}, outdated or variable search results \cite{DBLP:conf/naacl/LiPGGZ24/self-checker, khaliq2024ragar}, high computational costs, prompt sensitivity, and manual prompt design \cite{DBLP:conf/naacl/LiPGGZ24/self-checker}, and they may still miss fine‑grained details even when grounded \cite{tran2024leaf}. To mitigate these issues, current strategies include using IR functions (e.g., BM25) to distill relevant content \cite{tran2024leaf, quelle2024perils}, using multi‑agent \cite{liu2025bidev,leippold2025automated}, explicit decomposition and filtering \cite{zhao2024pacar, liu2025bidev,krishnamurthy2024yours}, feedback loops \cite{zhang2024reinforcement}, and refined RAG processes \cite{tran2024leaf, krishnamurthy2024yours}. Reliance solely on an LLM’s internal knowledge, whether via zero‑shot, few‑shot, or vanilla CoT prompting, is unreliable for fact‑checking and prone to substantial hallucination \cite{hu2024bad,zhang2023towards, tran2024leaf}, as zero‑shot CoT and vanilla CoT often succumb to memorization pitfalls \cite{hu2024bad, ma2025local}. Mitigating hallucination primarily involves incorporating external knowledge through ReAct, Search‑Augmented CoT, HiSS, and other RAG frameworks \cite{zhang2023towards, tran2024leaf, singhal2024evidence}.

In conclusion, while basic prompting strategies like zero-shot, few-shot, and vanilla CoT offer foundational ways to interact with LLMs for fact-checking, their accuracy and reliability are severely limited by the reliance on potentially flawed internal knowledge \cite{hu2024bad, zhang2023towards,tran2024leaf}. The most effective approaches, as highlighted by the sources, involve prompting strategies that explicitly integrate external knowledge retrieval through frameworks such as ReAct, HiSS, SELF-CHECKER, BiDeV, RAGAR, and PACAR \cite{DBLP:conf/naacl/LiPGGZ24/self-checker,zhang2023towards,liu2025bidev,quelle2024perils, singhal2024evidence}. These methods use prompts to guide LLM through processes that involve external data access, significantly improving accuracy and mitigating hallucination by grounding the model's responses in evidence \cite{DBLP:conf/naacl/LiPGGZ24/self-checker, zhang2023towards,singhal2024evidence}. 

Prompting is also used to generate explanations, although the utility and reliability of these explanations can vary \cite{zhang2023towards, si-etal-2024-large}. Limitations across strategies include sensitivity to prompt wording, computational cost, and the need for more robust and automated design methods \cite{DBLP:conf/naacl/LiPGGZ24/self-checker}.

\subsubsection{Fine-tuning Architectures for Optimizing Fact-checking Performance}

Fact-checking performance is primarily optimized through two primary approaches: (1) the development and fine-tuning of specific model architectures and (2) the design of advanced prompting strategies for LLMs. These methods are often combined within complex fact-checking pipelines.

\vspace{0.02\linewidth}\noindent\textbf{\textit{Fine-tuning smaller transformer models on synthetic data.}} This approach fine-tunes pre-trained transformer models on structured synthetic data, often combined with standard entailment datasets, to teach them nuanced fact-checking against grounding documents \cite{tang2024minicheck}. The strategy focuses on generating challenging training instances that help models verify atomic facts across multiple sentences, with models such as MiniCheck-FT5, RBTA, and DBTA outperforming larger LLMs such as GPT-4 in specific benchmarks like LLM-AGGREFACT \cite{tang2024minicheck}. Notably, MiniCheck-FT5 achieves a 4.3\% improvement over AlignScore using a significantly smaller dataset. Difficulty aggregating evidence and reasoning over multiple facts, the method mitigates these through targeted synthetic data and simple aggregation strategies like majority voting \cite{setty2024surprising}.

\vspace{0.02\linewidth}\noindent\textbf{\textit{Fine-tuning SLMs for task-specific performance.}} This method focuses on fine-tuning SLMs for task-specific applications like fake news detection. The strategy includes training SLM directly on the target dataset and using LLM-generated rationales via architectures like the ARG network, with a distilled version (ARG-D) for efficiency \cite{hu2024bad}. Fine-tuned BERT models have outperformed GPT-3.5 in fake news detection, and ARG/ARG-D exceed baseline methods that combine both SLM and LLM capabilities \cite{hu2024bad}. LLMs' difficulty in fully utilizing their reasoning for domain-specific tasks and their inability to fully replace SLMs in these contexts. Mitigation strategies include employing LLMs as rationale providers to guide SLMs, as well as exploring advanced prompting techniques and model combinations to achieve improved performance \cite{hu2024bad}. 

\vspace{0.02\linewidth}\noindent\textbf{\textit{Instruction Fine-tuned LLMs as Verifiers within a Pipeline Framework.}} This framework integrates an instruction-fine-tuned LLM as a verifier within a fact-checking pipeline, where the LLM evaluates claims based on contextually retrieved evidence. The process includes claim atomization using Mistral-7B, relevance-based evidence retrieval and re-ranking, and final inference by the LLM, producing interpretable credibility reports \cite{krishnamurthy2024yours}. The "Yours Truly" framework achieves a 94\% F1-score, significantly outperforming other systems, with the Claim Atomizer alone boosting performance from 64\% to 93\% \cite{krishnamurthy2024yours}.

\subsubsection{Domain-specific Training for Model Adaptation in Specialized Knowledge Areas}

Domain-specific adaptation and fine-tuning are highlighted as crucial strategies for enhancing the performance and reliability of models, particularly LLMs, in automated fact-checking within specialized knowledge areas. The need for such domain specificity arises from the observation that the factual accuracy and vulnerability of LLM to hallucinations can vary significantly between different domains. While general-purpose models may perform well in broad areas, models fine-tuned or adapted for specific domains, such as medicine or science, often demonstrate superior performance in those particular fields. Sources discuss the application of fact-checking techniques across various specialized domains, including medical/biomedical, political, scientific, law, general biographic, and news \cite{zhao2024pacar,  zhang2025dataset, tran2024leaf, qi2024sniffer, wang2025openfactcheck}.

Several approaches are explored to achieve domain adaptation in fact-checking systems. One involves fine-tuning smaller transformer models on target datasets specific to a domain or task within fact-checking, such as claim detection or veracity prediction, which has shown surprising efficacy and can outperform larger, general LLMs in specific contexts \cite{setty2024surprising, quelle2024perils}. Another approach directly involves fine-tuning LLMs themselves or using techniques like instruction-tuning on domain-relevant data or tasks \cite{tran2024leaf, qi2024sniffer, magomere2025claims, krishnamurthy2024yours, cheung2023factllama, quelle2024perils}. Frameworks like OpenFactCheck are proposed to allow users to customize fact-checkers for specific requirements, including domain specialization \cite{wang2025openfactcheck}.

\subsubsection{Comparative Summary and Trends}
From straightforward prompting to more complex fine-tuning and domain-specific training, the methods for using LLMs in fact-checking are changing. To inform general-purpose models such as GPT-3.5 and GPT-4, initial research focused on zero-shot and few-shot in-context learning. However, in fact-checking benchmarks, a notable trend indicates that smaller, optimized models can surprisingly beat larger LLMs, providing a more economical and effective solution \cite{setty2024surprising}.

Prompt engineering has also progressed beyond simple CoT. The rise of organized hierarchical prompting techniques, such as HiSS, which break down complicated claims into verifiable steps to reduce hallucinations and improve reasoning, is a definite trend \cite{zhang2023towards, vladika2025step}. Furthermore, domain-specific adaptation is becoming more and more important. Using specialized datasets and knowledge bases, models are being particularly trained or prompted for difficult domains such as news, climate science, and medicine to increase the accuracy when general knowledge is inadequate \cite{zhao2024medico,leippold2025automated}.

Integrating external domain-specific knowledge through methods like RAG or incorporating Knowledge Graphs is also a significant strategy to augment models with the necessary specialized information, sometimes in conjunction with fine-tuning or adaptation \cite{peng2023check,giarelis2024unified,cao2024multi, tran2024leaf, cheung2023factllama}. For example, the LEAF approach enhances medical question answering by integrating fact-checking results into RAG and using fact-checks for supervised fine-tuning \cite{tran2024leaf}. Evaluation datasets tailored to specific domains or types of claims, such as SciFact for scientific claims, Climate-FEVER for climate science, EXPERTQA spanning multiple fields, or medical QA datasets, are utilized to benchmark the effectiveness of these domain-adapted systems \cite{tang2024minicheck, zhao2024pacar, zhang2025dataset, tran2024leaf,leippold2025automated}. 

While interventions like fine-tuning and claim normalization improve robustness, performance can still degrade when evaluated significantly out-of-domain across different topics or platforms \cite{magomere2025claims}. The sources collectively underscore that effective fact-checking in specialized areas often requires models or frameworks specifically tailored to the domain's knowledge and nuances, moving beyond one-size-fits-all general approaches \cite{zhang2025dataset, tran2024leaf, magomere2025claims,wang2025openfactcheck}.

\subsection{Integration of RAG in Fact-Checking (RQ5)}

\begin{figure}[!ht]
    \centering
    \includegraphics[scale=0.09]{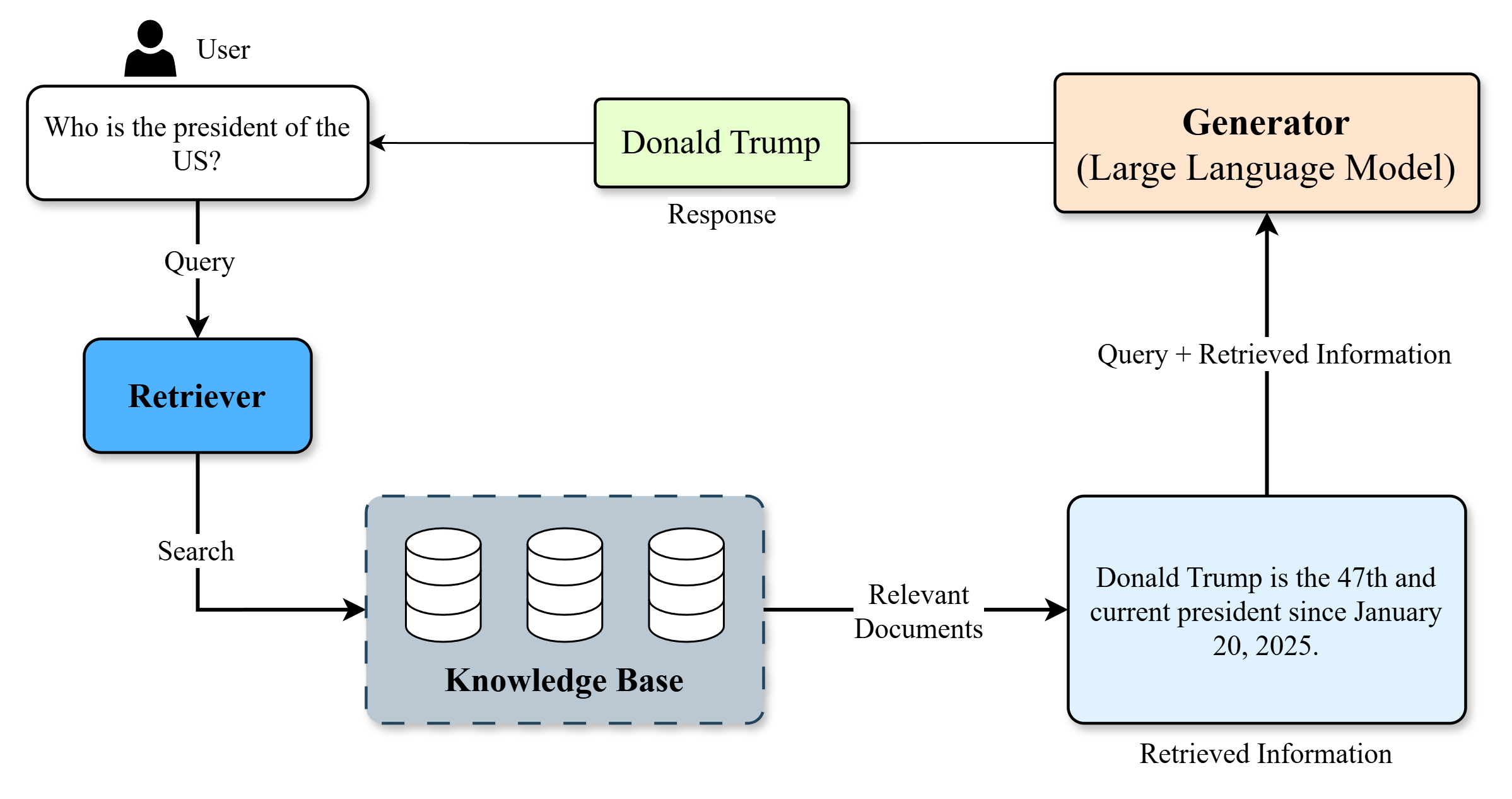}
    \caption{Workflow of a RAG system for factual question answering.}
    \label{fig:section - 4.5}
\end{figure}

RAG is a hybrid framework designed to enhance LLMs by integrating the retrieval of external knowledge into their generation process \cite{augenstein2024factuality,ding2025citations}. This approach is pivotal for grounding LLM outputs in external evidence, thereby mitigating common issues such as hallucination, the generation of factually incorrect or nonsensical information, and reliance on potentially outdated internal knowledge \cite{augenstein2024factuality, peng2023check, si-etal-2024-large}. Unlike methods solely based on fine-tuning the model's internal parameters, RAG employs external data sources, such as web search results or curated knowledge bases, to inform the LLM's responses during inference \cite{peng2023check,ding2025citations}. A significant advantage of RAG is its ability to produce responses that are factually accurate and reliable, and offer improved transparency and credibility by providing explicit citations to the \tb{external sources used \cite{quelle2024perils, ding2025citations,sairaj2025ensemble}. This capability allows} models to access and utilize current information, overcoming the limitations of knowledge cutoffs inherent in their training data \cite{khaliq2024ragar}. An overview of the workflow of a basic RAG-based system is presented in Figure \ref{fig:section - 4.5}.

In the domain of fact-checking, RAG systems play a crucial role in automating and improving the verification process \cite{quelle2024perils}. LLM agents empowered with RAG can phrase search queries based on claims, retrieve relevant external contextual data, and utilize this information to assess the veracity of statements \cite{quelle2024perils, ding2025citations}. This integration of external knowledge through RAG leads to enhanced accuracy in fact-checking, enabling the extraction of relevant evidence to support veracity predictions \cite{khaliq2024ragar,quelle2024perils, singhal2024evidence}. Approaches like Fact-Check-Then-RAG utilize the outcomes of fact-checking to refine the retrieval process itself and ensure that retrieved information specifically enhances factual accuracy \cite{tran2024leaf}. RAG also supports more complex scenarios, including multimodal fact-checking, where it is used to extract both textual and image content and retrieve external information for reasoning \cite{khaliq2024ragar,geng2024multimodal}. Furthermore, RAG-based systems can provide reasoned explanations for their verdicts, improving the interpretability of the fact-checking process \cite{zhao2024pacar,quelle2024perils}. Variations like FFRR utilize fine-grained feedback from the LLM to optimize retrieval policy based on how well documents support factual claims \cite{zhang2024reinforcement}. While some approaches explore reducing reliance on external retrieval by comprising the LLM's internal knowledge, RAG is generally considered essential for effective fake news detection \cite{geng2024multimodal, li2024large}.
\begin{figure*}[ht!]
    \centering
    \includegraphics[scale=0.15]{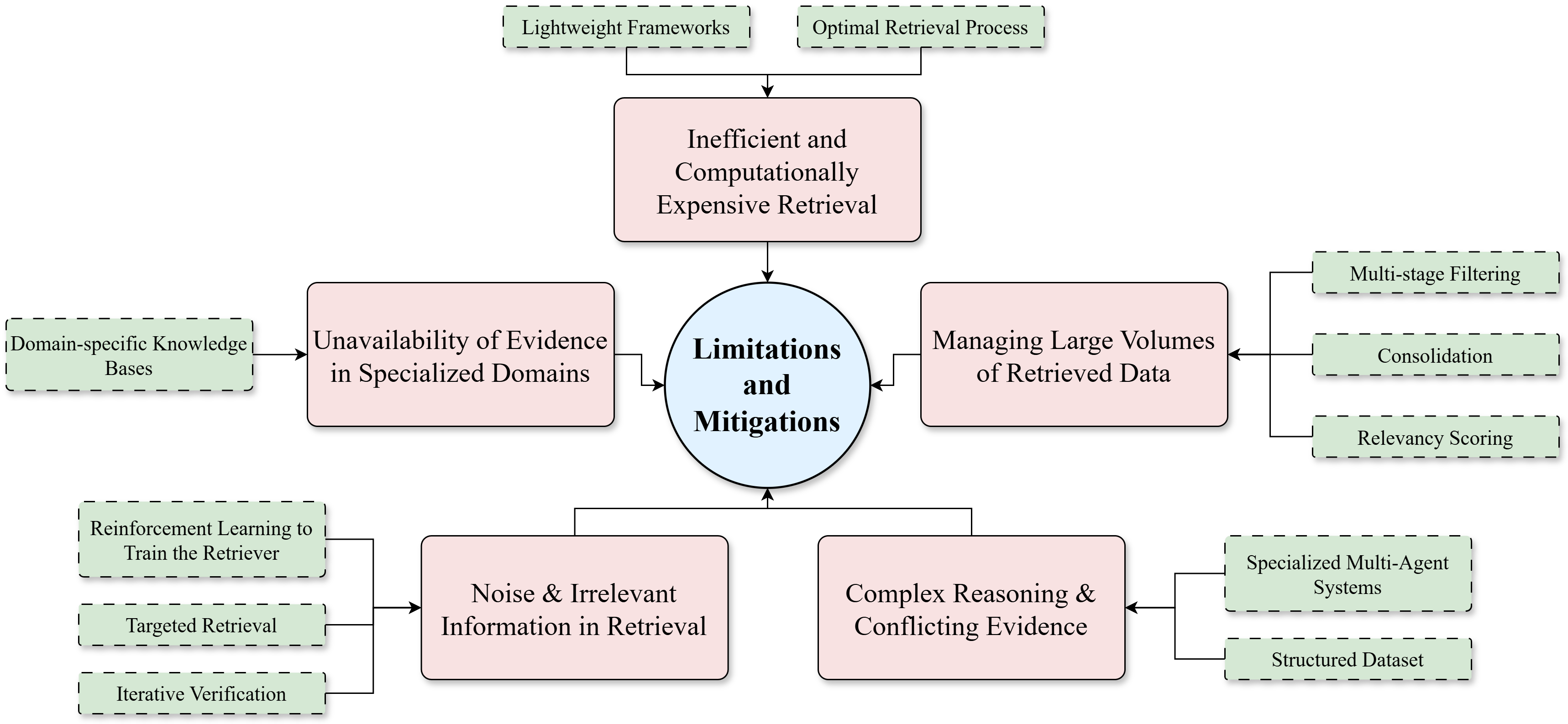}
    \caption{Limitations in RAG-based fact-checking and mitigation Strategies. The red boxes illustrate the limitations, while the green boxes represent the corresponding mitigation strategies.}
    \label{fig:section - 4.5_Limit_Mitigations}
\end{figure*}

Despite its advantages, implementing RAG, particularly in specialized domains, presents several challenges and limitations \cite{augenstein2024factuality, fadeeva2024fact}. A significant hurdle is the requirement for efficient and accurate retrieval of relevant evidence at scale, which can be a computational bottleneck \cite{augenstein2024factuality}. The effectiveness of RAG is inherently dependent on the assumption that pertinent information is readily available and accessible within the external knowledge sources used, such as search engines \cite{zhang2023towards}. This assumption may not hold for all information, especially in specialized or low-resource domains where relevant knowledge might be obscure, non-digitized, or exist in formats not easily indexed \cite{zhang2023towards}. Retrieving and processing large volumes of external data can also overwhelm the LLM's context window, necessitating sophisticated techniques for selecting and consolidating the most critical information \cite{peng2023check,quelle2024perils}. Standard RAG-based methods can inadvertently introduce noise or irrelevant information, potentially hindering the LLMs' performance rather than improving it \cite{tran2024leaf, zhang2024reinforcement}. In specialized fields like healthcare, general RAG models may struggle due to a lack of the nuanced understanding required for accurate fact-checking, highlighting the need for tailored or potentially fine-tuned approaches or integrating domain-specific knowledge bases \cite{giarelis2024unified,tran2024leaf}. Complex fact-checking tasks, such as interpreting conflicting evidence, handling claims with insufficient external information, or performing causal reasoning with fragmented information across documents, remain challenging for LLMs even with RAG \cite{liu2025bidev,singhal2024evidence, krishnamurthy2024yours}. Furthermore, the dynamic nature of information requires constant updates to external sources, and relying solely on search results can lead to diluted credibility if misinformation is widely reported \cite{li2024large}.

For inefficient and computationally expensive evidence retrieval, researchers have developed lightweight frameworks like Provenance \cite{sankararaman2024provenance}, which use compact and open-source NLI models for verification instead of LLMs. FIRE \cite{xie2024fire}, a framework that is designed for time and cost efficiency through an iterative process. Reinforcement retrieval models further show that a properly trained retriever does not add significant overhead during inference, as the costly feedback loop is only part of that training phase \cite{zhang2024reinforcement}. Due to the unavailability of evidence in specialized or low-resource domains, frameworks like LEAF \cite{tran2024leaf}, designed for the medical domain, use a specialized corpus such as MedRAG, which includes PubMed and textbooks, instead of relying solely on Google Search. The challenge of fact-checking in low-resource languages is mitigated by translating the claims into high-resource languages, such as English, in some literature \cite{quelle2024perils}. 

To prevent overwhelming the LLM’s context window and to efficiently manage large volumes of retrieved information, frameworks such as Provenance \cite{sankararaman2024provenance} employ a Relevancy Score in combination with TopK/TopP selection modules to filter the most critical evidence before it reaches the verifier. Similarly, LLM-AUGMENTER incorporates a "Knowledge Consolidator" that prunes irrelevant data and synthesizes the remaining evidence into concise reasoning chains \cite{zhang2024reinforcement}. Reinforcement Retrieval further demonstrates that limiting the number of documents, typically to the top three or four, can be more effective than including a larger set, which may introduce noise \cite{zhang2024reinforcement}. For mitigating noise and irrelevant information introduced by standard RAG methods, Reinforcement Retrieval addresses this directly using reinforcement learning to train the retriever; feedback from the LLM verifier acts as a reward signal, teaching the retriever to select more useful and factually relevant documents \cite{zhang2024reinforcement}. The LEAF framework uses a "Fact-Check-Then-RAG" approach, where an initial fact-check on the LLM's output is used to guide a more targeted and accurate retrieval process \cite{tran2024leaf}. 

Furthermore, to handle complex reasoning, conflicting evidence, and insufficient information, the PACAR framework employs a dynamic planner that can deploy tailored agents for tasks like numerical reasoning and entity disambiguation \cite{zhao2024pacar}. The LoCal framework is a multi-agent system specifically designed to handle the complexities of logical and causal fact-checking. For scenarios with conflicting or insufficient evidence \cite{schlichtkrull2023averitec}, the AVERITEC dataset was created with a "Conflicting Evidence" label and a structured question-answer format that can represent evidential disagreements, providing a basis for training models to better navigate such ambiguity \cite{schlichtkrull2023averitec}.

\subsubsection{Comparative Summary and Trends}
The incorporation of RAG, which has been a key strategy in LLM-based fact-checking, has established a trend away from dependence on static, internal knowledge. A complex claim is typically divided into verifiable sub-claims in the basic RAG pipeline \cite{zhang2023towards, zhao2024medico}, followed by the collection of relevant external evidence, its combination with a verifier model, and a final decision \cite{cheung2023factllama, bai2024large}. It is evident that in recent years, this linear pipeline has changed into more intelligent, dynamic, and effective systems. One significant advancement is the shift to agent-based and iterative frameworks. Systems like FIRE \cite{xie2024fire} and LLM-AUGMENTER \cite{peng2023check} can handle complex, multi-hop claims more successfully because they employ repeated cycles of retrieval and verification rather than a single retrieval step. Multi-agent systems like LoCal and PACAR \cite{zhao2024pacar}, which employ planning modules to dynamically choose which tools or reasoning processes to employ, further develop this. The growing complexity of the retrieval and verification procedure itself is another significant development. Frameworks such as Provenance \cite{sankararaman2024provenance} employ lightweight models to score and filter evidence for relevance rather than just providing an LLM's raw search results. As investigated in the Reinforcement Retrieval framework, there is also a shift toward optimizing the retriever for the downstream fact-checking job by employing strategies like reinforcement learning to transmit feedback from the verifier back to the retriever. An overview of the limitations and mitigation strategies is summarized in Figure \ref{fig:section - 4.5_Limit_Mitigations}.

\begin{table*}[ht!]
\begin{scriptsize}
\centering
\caption{Quantitative Synthesis of Key Findings on Fact-Checking Methodologies. }
\label{tab:quantitative_summary}
\begin{tabular}{p{4cm}|p{12cm}|>{\centering\arraybackslash}p{1cm}}
\midrule
\textbf{Claim \& Core Insight} & \textbf{Quantitative Evidence \& Context} & \textbf{\ Papers} \\
\midrule

\multirow{9}{4cm}{\RaggedRight Domain-Tuned SLMs Can Outperform Larger LLMs} &
A fine-tuned BERT model achieved a +9.0\% relative improvement in F1-score over the best-performing GPT-3.5-turbo configuration (76.5\% vs. 70.2\%) on the GossipCop dataset \cite{hu2024bad}. & \multirow{9}{*}{4} \\
& & \\
& A supervised ROBERTa-Base classifier outperformed zero-shot GPT-3.5-turbo by +9.6 absolute points in F1-macro on the FA-KES dataset (52.9\% vs. 43.3\%) \cite{leite2023detecting}. & \\
& & \\
& A fine-tuned BioBERT for clinical claims achieved 80.2\% accuracy on CliniFact, significantly outperforming both zero-shot (34.3\%) and fine-tuned (53.6\%) Llama3-70B \cite{zhang2025dataset}. &  \\
& & \\
& MiniCheck-FT5 model (770M parameters), fine-tuned on specially generated synthetic data, achieved an average Balanced Accuracy of 74.7\% on the LLM-AGGREFACT benchmark, which was statistically comparable to GPT-4's score of 75.3 GPT-4's score of 75.3\% \cite{tang2024minicheck}. & \\
\midrule

\multirow{5}{4cm}{\RaggedRight Advanced Prompting is Better than Standard CoT} &
The HiSS (Hierarchical Step-by-Step) method surpassed vanilla CoT by +9.5 absolute points in F1-score on the RAWFC dataset (53.9\% vs. 44.4\%). & \multirow{5}{*}{2} \\
& & \\
& On the LIAR dataset, HiSS outperformed vanilla CoT by +7.1 points in F1-score (31.3\% vs. 24.2\%). HiSS also outperformed the more advanced ReAct agent framework by +4.1 absolute points on RAWFC \cite{zhang2023towards}. & \\
& & \\
& At a detailed 5-class classification level (Level 2), the CLIMINATOR framework achieved an accuracy of 72.7\%. This was substantially better than the GPT-4o advocate alone, which only achieved 56.5\% accuracy on the same task\cite{leippold2025automated}. & \\
\midrule

\multirow{8}{4cm}{\RaggedRight RAG Provides Significant Performance Gains} &
Providing external context (RAG) to GPT-4 on the PolitiFact dataset increased its accuracy on non-ambiguous verdicts from 75\% to 89\%. The accuracy on "true" claims jumped by +13.62 absolute points \cite{quelle2024perils}. & \multirow{8}{*}{5} \\
& & \\
& The Fact-Check-Then-RAG method improved Llama 3 70B's accuracy on the PubMedQA dataset from 60.60\% to 73.60\% (+13.0 absolute points) by using fact-checking results to guide retrieval \cite{tran2024leaf}. & \\
& & \\
& An RAG pipeline using Mixtral achieved a 0.780 F1-score on the 'Refuted' class on the Averitec development set \cite{singhal2024evidence}. & \\
& & \\
& On the RAWFC dataset, the fine-tuned FactLLaMA model with external knowledge achieved a Macro-F1 score of 0.5565. This significantly outperformed the same model fine-tuned without external knowledge, which only scored 0.5376 \cite{cheung2023factllama} & \\
& & \\
& On the LIAR-RAW dataset, "Direct Prompting" of the LLM gave an F1 score of 27.0\%. Simply adding a frozen (non-optimized) retriever in the FFRR-frozen setup increased the F1 score to 30.1\%, demonstrating the inherent benefit of RAG. \cite{zhang2024reinforcement} & \\
\midrule

\multirow{11}{4cm}{\RaggedRight Hybrid and Multi-Agent Systems are More Effective} &
Compared to strong LLM baselines like Flan-T5 and ChatGPT, the LoCal multi-agent system showed an average performance improvement of up to 7.75\% in the gold evidence setting and up to 6.17\% in the open book setting \cite{ma2025local}. & \multirow{11}{*}{4} \\
& & \\
& The hybrid SLM+LLM ARG network improved F1-score over its BERT-only baseline by +3.1 absolute points on the Weibo21 dataset (78.4\% vs. 75.3\%) \cite{hu2024bad}. & \\
& & \\
& The PACAR framework, with specialized agents, outperformed a general ChatGPT baseline by +16.9 absolute points on HOVER 4-hop claims (72.61\% vs. 55.72\%) \cite{zhao2024pacar}. & \\
& & \\
& The FACT-AUDIT adaptive multi-agent framework demonstrated superior evaluation robustness over static, single-agent pipelines by dynamically assigning roles \cite{lin2025fact}. & \\
\midrule

\multirow{11}{4cm}{\RaggedRight Automated Feedback Mechanisms Reduce Hallucinations} &
The LLM-AUGMENTER system, using a BM25 knowledge consolidator and automated feedback, improved the KF1 score to 37.41 over the GPT KF1 score of 31.33 \cite{peng2023check}. & \multirow{11}{*}{4} \\
& & \\
& The Self-Checker framework improved label accuracy on the BINGCHECK dataset from 21.0\% (ReAct baseline) to 63.4\% \cite{DBLP:conf/naacl/LiPGGZ24/self-checker}. & \\
& & \\
& Medico's multi-source evidence fusion and correction loop improved hallucination detection F1-score by +34.4 points over its baseline on the HaluEval dataset \cite{zhao2024medico}. & \\
& & \\
& The Visual Fact Checker uses object detection and VQA models as automated "tools" to verify and correct initial caption proposals, significantly reducing hallucinations in detailed image captions \cite{ge2024visual}. & \\
\midrule

\multirow{8}{4cm}{\RaggedRight Fine-tuning on Synthetic Data Boosts Performance} &
The MiniCheck-FT5 model, trained on generated synthetic data, achieved a Balanced Accuracy of 74.7\% on the LLM-AGGREFACT benchmark, a +4.3 absolute point improvement over the previous state-of-the-art. An ablation study showed that removing this synthetic data caused the model's performance to drop by -14.8 absolute points \cite{tang2024minicheck}. & \multirow{8}{*}{2} \\
& & \\
& FACT-GPT was trained on a synthetic dataset of contradicting, entailing, or neutral claims generated by GPT-4, which enabled a smaller, specialized LLM to match the claim-matching accuracy of larger models \cite{choi2024fact}. & \\
\midrule
\end{tabular}
\end{scriptsize}
\end{table*}

\section{Discussion}
\label{discussion}

Our review explored the rapid adoption of LLMs in the complex task of automated fact-checking. By examining a wide range of current studies, we have mapped out how we measure their success, the persistent problem of models generating false information (hallucinations), and the essential role and limitations of the datasets they rely on. We also looked into methods for improving LLMs, from prompt engineering and fine-tuning to the increasingly vital use of RAG. The research landscape reveals a field that is buzzing with innovation and showing great promise. It simultaneously highlights significant and complex hurdles that remain. If LLMs are to become truly reliable tools in the global effort against misinformation, these challenges demand ongoing and rigorous investigation \cite{augenstein2024factuality, quelle2024perils}.

\vspace{0.02\linewidth}\noindent\textbf{\textit{Evaluation metrics.}} In the evaluation metrics domain (RQ1) for LLM-based fact-checking, we can see the clear transition in classification scores towards more sophisticated, holistic, and context-aware frameworks. The rise of rigorous benchmarks such as LLM-AGGREFACT \cite{tang2024minicheck} and the AVERITEC dataset \cite{schlichtkrull2023averitec}, along with new centralized tools such as OpenFactCheck \cite{wang2025openfactcheck}, marks a major step towards creating shared ways to measure how accurate LLMs are. However, even with this progress, many important issues remain unresolved. Furthermore, making models tough enough to handle misinformation that adapts, often as an adversary, including claims with subtle edits or those that change over time, remains a major hurdle \cite{magomere2025claims}. The lack of strong methods for clear explanation and adaptation to new types of risks makes it harder for everyday users to judge whether the answers from LLMs are true. This shortfall also weakens the usefulness of these models in fast-changing real-life situations.

Two key challenges are now drawing more attention: teaching models how to check their work and catch their own mistakes \cite{kamoi2024evaluating}, and finding reliable ways to measure how closely their responses match the source material \cite{jing2024scale}. Despite advances in automated and LLM-driven assessments, Human Evaluation remains essential for nuanced aspects like explanation quality and contextual appropriateness \cite{deverna2024fact}, although it is resource-intensive. In general, while progress is evident, significant lacunae persist. There is a pressing need for standardized metrics that robustly assess the quality of LLM-generated explanations \cite{qi2024sniffer}, the resilience of the model against evolving misinformation \cite{magomere2025claims}, and the logical integrity of reasoning pathways \cite{ghosh2024logical}. These developments underscore an urgent and ongoing need for metrics that can holistically evaluate not only the veracity of claims, but also the provenance of supporting evidence \cite{sankararaman2024provenance} and the logical integrity of the LLM's reasoning process.

\vspace{0.02\linewidth}\noindent\textbf{\textit{Hallucination in LLMs.}} The tendency of LLMs to hallucinate (RQ2), that is, to generate outputs that are linguistically fluent and coherent yet factually misleading or entirely unsubstantiated, remains a significant barrier to their trustworthy deployment in sensitive, high-stakes applications such as fact-checking \cite{augenstein2024factuality, zhang2023towards}. RAG has become a foundational mitigation strategy, designed to ground LLM responses in verifiable external knowledge and reduce the models' reliance on their internal, and potentially flawed or outdated, parametric knowledge \cite{peng2023check, cheung2023factllama, quelle2024perils, bai2024large}. New efforts to boost how truthful language models are include adding systems that automatically give feedback, helping refine answers over multiple tries \cite{peng2023check}. Tools like Self-Checker are also being built. These are smart correction modules that let the model review and fix its own mistakes \cite{DBLP:conf/naacl/LiPGGZ24/self-checker}. On top of that, researchers are testing ways to bring together evidence from several sources, with models like MEDICO showing how this kind of fusion can work \cite{zhao2024medico}. This persistence of hallucinations is partly due to the inherent "black box" nature of current LLMs, the difficulty in comprehensively modeling nuanced world knowledge, and the escalating sophistication of adversarial attacks designed to exploit model vulnerabilities \cite{jing2024scale}.

\vspace{0.02\linewidth}\noindent\textbf{\textit{Datasets for Fact-Checking.}} The datasets (RQ3) utilized for the training, fine-tuning, and rigorous evaluation of fact-checking LLMs are pivotal to their ultimate performance and generalizability. While foundational datasets such as FEVER \cite{thorne2018fever} have been instrumental in catalyzing early research, the field is increasingly recognizing the necessity for more specialized, challenging, and contextually rich benchmarks. Illustrative examples include CliniFact, which is tailored for claims within the domain of clinical research \cite{zhang2025dataset}, AVERITEC, with its distinct emphasis on real-world claims that necessitate web-based evidence retrieval \cite{schlichtkrull2023averitec}, and BINGCHECK, specifically designed for assessing the factuality of LLM-generated text \cite{DBLP:conf/naacl/LiPGGZ24/self-checker}. The intrinsic characteristics of these datasets, including their composition, scale, annotation quality, topical diversity, and potential inherent biases (e.g., political, cultural, or temporal) profoundly influence model performance, the ability of models to generalize to unseen domains or claim structures, and, consequently, the perceived effectiveness of various fact-checking methodologies \cite{chatrath2024fact, yang2024fact}. The development and meticulous curation of robust multilingual datasets \cite{singhal2024multilingual} also represent a critical frontier to advance the global applicability and equity of LLM-based fact-checking technologies.

\vspace{0.02\linewidth}\noindent\textbf{\textit{Optimization Strategies and Domain-specific Training.}} Prompt engineering, fine-tuning strategies, and domain-specific training (RQ4) have been shown to significantly modulate the efficacy of LLMs in complex fact-checking tasks. Advanced prompting techniques, such as hierarchical step-by-step verification methods \cite{zhang2023towards} or structured reasoning frameworks like PACAR \cite{zhao2024pacar} and BiDeV \cite{liu2025bidev}, frequently demonstrate superior outcomes when compared to simpler, more direct prompting approaches. The application of zero-shot and few-shot learning paradigms is also being actively explored for a range of related tasks, including claim matching \cite{pisarevskaya2025zero, choi2024automated,choi2024fact}, indicating a strong potential for efficient adaptation of LLMs with limited task-specific data. Furthermore, the surprising efficacy of smaller, specifically fine-tuned transformer models in certain fact-checking contexts \cite{setty2024surprising} compellingly suggests that model scale is not the sole, nor always the primary, determinant of performance. This finding challenges the prevailing "bigger is better" narrative in LLM development and highlights the potential for more resource-efficient, specialized models to achieve competitive or even superior performance in targeted fact-checking applications, particularly when data and computational budgets are constrained. Domain-specific adaptations, which may involve the utilization of LLM-predicted credibility signals \cite{leite2023detecting} or the development of highly specialized systems for critical domains such as medicine \cite{vladika2025step}, are proving essential for achieving the nuanced understanding and reliable fact verification required in these contexts.

\vspace{0.02\linewidth}\noindent\textbf{\textit{The Integration of RAG.}} Incorporating LLM with RAG (RQ5) is increasingly recognized as a central and indispensable strategy for enhancing the factuality of LLM outputs. This is achieved by providing models with dynamic access to external, often real-time, knowledge sources, thereby augmenting their inherent capabilities \cite{peng2023check, cheung2023factllama, quelle2024perils, bai2024large}. A framework such as FIRE \cite{xie2024fire} is being developed to optimize iterative retrieval and verification processes, aiming for greater efficiency and accuracy. Other lines of research explore the application of reinforcement learning techniques to refine and optimize retrieval strategies \cite{zhang2023towards}. Nevertheless, significant challenges remain within the RAG pipeline. These include the efficient and precise retrieval of truly relevant evidence from vast, heterogeneous, and often noisy information spaces; the effective fusion of information derived from multiple, potentially contradictory, sources \cite{zhao2024pacar}. Extending RAG to effectively handle multimodal inputs \cite{khaliq2024ragar,geng2024multimodal} and optimizing the timing and contextual relevance of information recommended by RAG-empowered agents \cite{sakurai2024llm} remain active and critical areas of ongoing investigation.

\section{Open issues and challenges}
\label{open_issues_challenges}

While LLMs have advanced fact-checking capabilities, several core challenges remain. These include mismatches between fluency and truth, domain limitations, and weak reasoning integration. 
\begin{table*}[ht!]
\begin{scriptsize}
\centering
\caption{Identified issues in LLM-based fact-checking and their implications. This includes observed behaviors, underlying causes, and the significance of each issue for reliable deployment.}
\label{tab:open_issues_table}
\begin{tabular}{p{3cm}|p{3.3cm}p{4.2cm}p{5.7cm}p{0.7cm}}
\midrule
\textbf{Issue/Challenge} & \textbf{Observed Behavior} & \textbf{Implication} & \textbf{Why does it matter?} & \textbf{Ref.} \\
\midrule
Mismatch Between Output Quality and Factual Accuracy 
& Models write very fluent and convincing text. 
& High-quality language does not mean the facts are correct.
& 1. Current evaluation methods favor “sounding good” over “being accurate.” \newline
2. Models may get high scores for responses that look right but contain factual errors.
& \cite{quelle2024perils} \cite{bang2023multitask} \cite{guerreiro2023hallucinations} \\
\midrule
Limited Relevance Across Domains and Languages 
& Models perform well on simple, synthetic, or English-only datasets. 
& Struggle with real-world, complex, or multilingual data.
& 1. Fact-checking needs to work across many subjects, topics, and languages. \newline
2. Limited data variety in training/testing leads to poor generalization. 
& \cite{huang2025survey} \cite{choi2024fact} \\
\midrule
Challenges in Retrieval and Prompting Mechanisms 
& Use of RAG brings in external evidence
& 1. Retrieval is often imperfect (brings irrelevant or noisy info). \newline
2. Advanced prompting (CoT, multi-agent) still leads to error cascades.
& 1. Fact-checking relies on reliable evidence retrieval and reasoning chains. \newline
2. Weaknesses here mean incorrect or unsupported conclusions. 
& \cite{zhao2024pacar} \cite{fadeeva2024fact} \\
\midrule
Lack of Integration with Symbolic or Structured Reasoning 
& Current LLMs rely mostly on pattern recognition, not logic.
& 1. Little integration with logic/symbolic systems. \newline
2. Models can not follow strict, logical reasoning pipelines.
& 1. Symbolic reasoning would make fact-checking more robust and explainable. \newline
2. Lack of it = less trustworthy and harder-to-monitor systems. 
& \cite{leippold2025automated} \cite{sakurai2024llm} \\
\midrule
\end{tabular}
\end{scriptsize}
\end{table*}

Despite the advancement in fact-checking using LLMs, one common challenge remains the gap between model-generated responses' linguistic quality and factual accuracy. Today's criteria of evaluation are inclined to appreciate language or text overlap with references, which can overlook basic fact errors. It guides to feedbacks that look questionable but sound good and is inaccurate. These responses can receive arbitrarily high scores, which makes the system seem more accurate than it is \cite{quelle2024perils, bang2023multitask, guerreiro2023hallucinations}.

Existing models often execute much better on small synthetic datasets, but often fail to generalize well to real-world circumstances. The reason lies in the limited datasets with low complexity, variation in topics, and multilingual texts. Consequently, the models do not handle the variability and nuance of real-world data across languages and topics, which implies the need for more realistic and broad training and test corpora \cite{huang2025survey, choi2024fact}. Additionally, Siino et al. added that transformer-based models generally perform the worst in terms of standard deviation \cite{siino2022fake}. However, RAG techniques have a stronger evidence-based foundation in LLM, but retrieval is imperfect. Fact-correctness is often weakened by the quality of the retrieved information, noisy or irrelevant documents, and when there is insufficient context available. Similarly, advanced prompting techniques, such as CoT reasoning or multi-agent collaborations, still must be precisely fine-tuned, but they also remain vulnerable to cascading errors in output generation \cite{zhao2024pacar,  fadeeva2024fact, language_action_review_large}.

One of the areas with great potential that is still underexplored lies in the integration of LLMs with symbolic reasoning or structured logic-based systems. These systems can improve interpretability and fact resilience in fact-checking pipelines. But work in this area is still in an early stage, and significant effort is needed to design scalable, proper architectures that can balance the respective strengths of neural and symbolic methods \cite{sakurai2024llm, leippold2025automated}. Table \ref{tab:open_issues_table} provides an organized summary of the existing issues and challenges discussed in this section.

% =================================================================

\section{Critical analysis of future research agendas}
\label{critical_analysis_future}
LLMs show great promise in automating fact-checking, but their use also highlights a range of ongoing problems that still need attention. In the future, research must take a thoughtful and future-focused approach. If these models are to become trustworthy, accurate, and ethically sound tools in fact-checking, the gaps in today's research, for example, the unreliability of retrieved evidence, the difficulty in verifying complex claims requiring multistep reasoning, and the challenge of mitigating factual hallucinations, need to be tackled head-on. What follows is a breakdown of key areas where further study could really make a difference, each pointing to where progress is most needed.

\begin{table*}[!ht]
\centering
\caption{Identified gaps from the proposed RQs, and future research paths in fact-checking with LLMs.}
\label{tab:future_research_agendas}
\begin{scriptsize}
 
\begin{tabular}{c|p{3.5cm}|p{13cm}}
\toprule
\textbf{RQs} & \textbf{Research Gaps} & \textbf{Potential Research Paths} \\
\midrule
\textbf{RQ1} & Evaluation and Benchmarking Challenges &
1. Develop a new evaluation matrix that not only limits itself to overlapping or semantic scores but also incorporates the factual correctness and the reasoning capabilities of LLMs, along with real-world dynamics and practicality. \newline
2. Establish robust metrics and methodologies for evaluating the human-computer interaction aspects of fact-checking systems in terms of clarity, actionability, and persuasiveness of explanations. \newline
3. Create evaluation frameworks that can assess a fact-checking system's ability to handle temporally sensitive claims, outdated evidence, and the "freshness" of information. \\
\midrule
\textbf{RQ2} & Trust and Reliability &
1. Develop automatic detection and correction methods for hallucinated LLM outputs, including uncertainty quantification. \newline
2. Optimize retrieval strategies with reinforcement learning and iterative verification to improve efficiency and accuracy. \newline
3. Investigate effective formats of AI-generated fact checks to enhance human trust through transparent explanations and evidence. \\
\midrule
\textbf{RQ3} & Limited Realistic, Complex, Multilingual Datasets &
1. Develop more realistic, complex, dynamic, domain-specific, multilingual fact-checking datasets with high-quality evidence for evaluation and fine-tuning. \newline
2. Develop innovative and efficient data creation and annotation methodologies. \newline
3. Develop more robust weak supervision, semi-supervised, or active learning techniques to reduce reliance on fully manual annotation. \newline
4. Design systems and protocols for continuous data collection and dataset updates to reflect the real-time nature of information and misinformation. \\
\midrule
\textbf{RQ4} & Prompt Sensitivity and Adaptation Challenges &
1. Design and evaluate prompting methodologies that explicitly enforce and enable verification of evidence-grounded and faithful reasoning. \newline
2. Develop adaptive, model-aware prompting frameworks that automatically generate and refine prompts and in-context examples to ensure robustness against variations. \newline
3. Develop continual learning strategies for fine-tuned and domain-specific models to allow them to adapt to new information; investigate meta-learning or adaptive techniques for prompt optimization. \newline
4. Develop prompting and fine-tuning methodologies that explicitly optimize for generating controllable, verifiable, and evidence-grounded reasoning and explanations. \\
\midrule
\textbf{RQ5} & Efficient Explainable Retrieval &
1. Design systems where LLMs can iteratively refine queries, explore multiple information angles, or retrieve evidence for decomposed sub-claims to build a more comprehensive evidence base. \newline
2. Advanced agent-based RAG systems where an LLM (or multiple specialized LLM agents) can plan a sequence of reasoning and retrieval steps. \newline
3. Design efficient RAG architectures that minimize computational overhead through optimized context chunking, selective retrieval, and reusable memory. \\
\bottomrule
\end{tabular}
   
\end{scriptsize}
\end{table*}

% \subsection{Advancing Evaluation Frameworks Beyond Conventional Metrics}
\vspace{0.02\linewidth}\noindent\textbf{\textit{Evaluation Framework Advancement.}} A fundamental imperative lies in transcending current evaluation metrics, which often inadequately capture the nuanced performance characteristics of LLMs in fact-checking tasks \cite{augenstein2024factuality,wang2025openfactcheck}. Future research must prioritize the development of sophisticated, multi-dimensional frameworks. This includes establishing standardized metrics for explainability and interpretability that demonstrably correlate with human cognitive trust and facilitate diagnostic understanding of model failures \cite{qi2024sniffer, vladika2025step, ding2025citations, abian2024auto, asppst, sheng2025uncertainty, mathew2025recent, cheung2025effectiveness}. Furthermore, the creation of dynamic evaluation suites (testing systems that can actively change and adapt over time, rather than relying on fixed, unchanging datasets) to rigorously test resilience against evolving misinformation tactics and sophisticated adversarial attacks is paramount \cite{magomere2025claims}, moving beyond static benchmarks. Currently, the development of precise metrics for fine-grained faithfulness and verifiable provenance tracking \cite{sankararaman2024provenance, jing2024scale} is crucial to ensure that LLM outputs are not merely plausible but demonstrably grounded in credible evidence.

% \subsection{Proactive Hallucination Mitigation and Enhanced Factual Grounding}
\vspace{0.02\linewidth}\noindent\textbf{\textit{Factual Hallucination Mitigation.}} The mitigation of LLM-generated hallucinations requires a strategic shift from reactive correction to proactive prevention mechanisms embedded within LLM architectures and training paradigms \cite{zhao2024pacar, zhao2024medico}. Currently, optimizing RAG systems to effectively navigate complex and noisy information environments \cite{zhang2023towards, liu2025bidev} and enabling dynamic and reliable knowledge updates within LLMs \cite{augenstein2024factuality} are critical to ensure accurate and consistent factual grounding. Without these, RAG systems risk amplifying, rather than rectifying, inaccuracies.

% \subsection{Enhancing Logical Consistency, Reasoning, and Calibrated Trust}
\vspace{0.02\linewidth}\noindent\textbf{\textit{Logical Consistency, Reasoning, and Calibrated Trust.}} The ultimate efficacy of LLMs in fact-checking is critically dependent upon their capacity for robust logical reasoning and their ability to engender warranted user trust. Future endeavors must explore formal verification methods (mathematically and logically ensuring that the model's reasoning process is sound and its conclusions are consistent) to enhance the logical consistency of LLM outputs \cite{ghosh2024logical} and significantly deepen their reasoning capabilities for complex, multihop, and inferential claims \cite{zhao2024medico}. The degree to which users believe the fact-checks offered by LLMs must also be thoroughly investigated. The goal of this study should be to appropriately "calibrate" or modify that trust to a suitable degree. The purpose is to encourage people to critically evaluate the information provided rather than relying too much on these automatic checks \cite{ding2025citations, deverna2024fact}

% \subsection{Expanding Frontiers: Multimodality and Multilinguality}
\vspace{0.02\linewidth}\noindent\textbf{\textit{Multimodality and Multilinguality.}} Given that misinformation transcends unimodal (ie, English) texts, even though some studies have tried to incorporate multimodal and multilingual fact-checking \cite{singhal2024multilingual, ge2024visual, quelle2024perils, geng2024multimodal}, there is still a need for a significant expansion of LLM fact-checking capabilities. The development of robust multimodal systems capable of verifying claims that integrate textual, visual, and auditory information, and detecting sophisticated cross-modal manipulations is a key frontier \cite{qi2024sniffer, ge2024visual}. Equally vital are dedicated efforts to develop and evaluate information and effective and equitable fact-checking in a diverse spectrum of languages, with particular attention to resource-scarce linguistic contexts \cite{singhal2024multilingual, singhal2024evidence}.

Table \ref{tab:future_research_agendas} provides an overview of the identified gaps and future research agendas.

\section{Conclusion}
\label{conclusion}

The inclusion of LLMs in automated fact-checking is changing the field. These models are reshaping how we process and verify the overwhelming amount of information we face online. Our work lays out the current research in this space, drawing attention to five critical areas: how LLMs are evaluated, the problem of hallucinations (where models produce false information), the importance of data sources, various ways of improving performance, and the growing use of RAG. Furthermore, in addition to extensive textual fact-checking, our review also acknowledges assessment of multimodal and multilingual dimensions, recognized as core areas of contemporary research. The findings highlight the fact that while this field is moving quickly and holds considerable promise, there are significant challenges that still need careful attention if we want these systems to work reliably.

One of the major achievements of this study is that it pulls together a broad snapshot of what is going on right now. It highlights a tricky balance: on the one hand, LLMs have the potential to improve the speed and quality of fact-checking. However, these systems are still limited, and fixing those limits is going to require a lot of effort from researchers. It is essential to examine the practical utility of employing LLMs across all tasks, as smaller, task-specific models can often outperform them when properly optimized. From prior studies, one thing becomes clear: we need better tools to judge these systems. Since most existing tools only focus on whether a fact is right or wrong, there is now a growing need for tools that also look at how clearly the model explains things, whether it sticks to the logic, how well it shows where its answers came from, and how tough it is against tricky questions or misleading setups.  From our review, it is evident that it is still difficult to ensure that LLMs always stick to real facts, especially when the situation is new or tricky. This improvement will require fundamental changes in the way these models are built and trained.

Although this review offers a comprehensive synthesis of the existing literature, its scope is inherently constrained by the rapid velocity of technological innovation within the LLM domain. Consequently, emerging preprint findings, advancements in proprietary models, and developing best practices may not be fully encapsulated. Looking ahead, future research must prioritize the development of more sophisticated, robust, and universally standardized evaluation benchmarks. Such benchmarks are urgently needed to assess the factual accuracy, logical coherence, soundness of LLM reasoning, quality, utility, and persuasiveness of generated explanations, as well as the models' resilience to a wide array of adversarial attacks and evolving misinformation tactics.

The review presents key insights for developers, policymakers, and all stakeholders who rely on online information. It underscores the need to address significant shortcomings in the development and evaluation of LLMs to enable their effective use as reliable fact-checking tools. That means paying close attention to how fair and balanced the data are, learning how to handle different languages and types of media better, and setting up strong rules and protections to make sure these systems are used ethically and do not cause harm.

% Our review makes a powerful case: the future of fact-checking is not about handing everything over to machines. It is about building strong teams—human experts working side by side with smart machines. LLMs are tools, not magic fixes, and their true value will only show when we put in the effort to use them wisely. This path would not be easy. It is going to take time, debate, and deep thinking. But the insights in this review give us a solid foundation to work from—one that could help build a more informed, thoughtful, and resilient way of dealing with misinformation.

\section*{Declarations}
\noindent
\textbf{Conflict of Interests:} On behalf of all authors, the corresponding author states that there is no conflict of interest.
\textbf{Funding:} No external funding is available for this research.\\
\textbf{Data Availability Statement:} Not Applicable.\\
\textbf{Ethics Approval and Consent to Participate}. Not Applicable.
\\
\textbf{Informed Consents:} Not Applicable.\\
\textbf{Author Contributions:}
\textit{Conceptualization and Methodology:} Subhey Sadi Rahman, Md. Adnanul Islam, Md. Mahbub Alam, Musarrat Zeba, Mohaimenul Azam Khan Raiaan;

\textit{Resources and Literature Review:} Subhey Sadi Rahman, Md. Adnanul Islam, Md. Mahbub Alam, Musarrat Zeba;

\textit{Writing – Original Draft Preparation:} Subhey Sadi Rahman, Md. Adnanul Islam, Md. Mahbub Alam, Musarrat Zeba, Md. Abdur Rahman, Sadia Sultana Chowa, Mohaimenul Azam Khan Raiaan;

\textit{Validation: }Sami Azam, Sadia Sultana Chowa, Mohaimenul Azam Khan Raiaan, Md. Abdur Rahman;

\textit{Formal Analysis:} Md. Abdur Rahman, Sami Azam;

\textit{Writing – Reviewing and Finalization:} Sami Azam, Mohaimenul Azam Khan Raiaan;

\textit{Project Supervision:} Sami Azam, Mohaimenul Azam Khan Raiaan;

\textit{Project Administration:} Sami Azam, Mohaimenul Azam Khan Raiaan.

\end{document}